\documentclass[10pt,twocolumn,letterpaper]{article}

\usepackage[pagenumbers]{cvpr} 

\usepackage{graphicx}
\graphicspath{{./figs/}}
\usepackage{amsmath}
\usepackage{amssymb}
\usepackage{booktabs}

\usepackage{enumerate}

\usepackage[pagebackref,breaklinks,colorlinks]{hyperref}

\usepackage{algorithm} 
\usepackage{algpseudocode} 
\usepackage{multirow}

\usepackage[capitalize]{cleveref}
\crefname{section}{Sec.}{Secs.}
\Crefname{section}{Section}{Sections}
\Crefname{table}{Table}{Tables}
\crefname{table}{Tab.}{Tabs.}

\def\cvprPaperID{2373} 
\def\confName{CVPR}
\def\confYear{2022}

\newcommand{\eqpref}{Eq.~}
\newcommand{\secpref}{Sec.~}

\begin{document}
	
	\title{RIO: Rotation-equivariance supervised learning of robust inertial odometry}
	
	\author{Xiya Cao\textsuperscript{1}\thanks{\textsuperscript{1} denotes equal contribution.}
		\qquad
		Caifa Zhou\textsuperscript{1}
		\qquad
		Dandan Zeng
		\qquad
		Yongliang Wang\\
		Riemann lab, 2012 Laboratories, Huawei Technologies Co.~Ltd\\
	}
	
	\maketitle
	
	\begin{abstract}
		
		This paper introduces rotation-equivariance as a self-supervisor to train inertial odometry models. We demonstrate that the self-supervised scheme provides a powerful supervisory signal at training phase as well as at inference stage. It reduces the reliance on massive amounts of labeled data for training a robust model and makes it possible to update the model using various unlabeled data. Further, we propose adaptive Test-Time Training (TTT) based on uncertainty estimations in order to enhance the generalizability of the inertial odometry to various unseen data. We show in experiments that the Rotation-equivariance-supervised Inertial Odometry (RIO) trained with 30\% data achieves on par performance with a model trained with the whole database. Adaptive TTT improves models performance in all cases and makes more than 25\% improvements under several scenarios.
		
	\end{abstract}
	

	\section{Introduction}

Accurate and robust localization with low-cost Inertial Measurement Units (IMUs) is an ideal solution to a wide range of applications from augmented reality\cite{zhou2020integrated} to indoor positioning services\cite{wang2016indoor, zhou2021mining}. An IMU usually consists of accelerometers and gyroscopes, sometimes magnetometers and can sample linear acceleration, angular velocity and magnetic flux density in an energy-efficient way. It can be light-weight and pretty cheap that many mobile devices like smartphones and VR headsets are instrumented with it. In many scenarios such as indoor or underground where global navigation satellite system is not available, ubiquitous IMU is a promising signal source, which can provide reliable and continuous location service. Unlike Visual-Inertial Odometry (VIO)\cite{forster2016manifold} that is sensitive to surroundings and cannot work under extreme lighting, IMU-only inertial odometry is more desired and possible to perform accurate and robust localization every time and everywhere\cite{harle2013survey, liu2020tlio}.

\begin{figure}
	\centering
	\includegraphics[width=8.9cm]{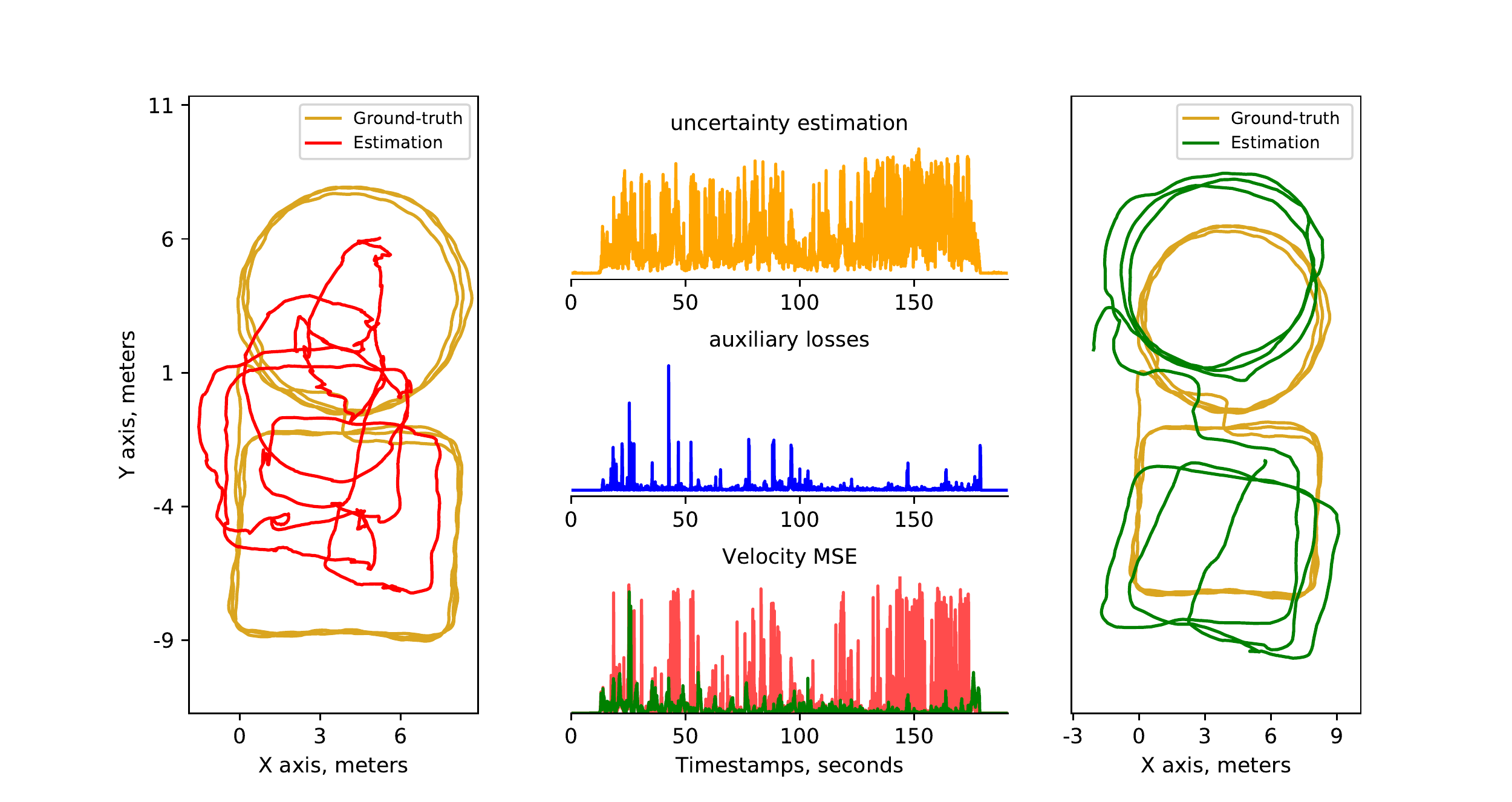}
	\vspace{-4ex}
	\caption{An example trajectory estimation improved by RIO. On the left, we show original model performance before rotation-equivariance supervised learning and right is the result of RIO. In the middle, we show uncertainty estimation (orange), auxiliary losses (blue) computed by the self-supervised task, original estimated velocity MSE (red) and updated model velocity MSE (green).}
	\label{fig:main_traj}
\end{figure}

Recent advances of data-driven approaches (e.g., IONet\cite{chen2018ionet}, RoNIN\cite{herath2020ronin}, TLIO\cite{liu2020tlio}) based on machine learning and deep learning have pushed the limit of traditional inertial odometry\cite{park2010zero, ho2016step} with the help of kinematic model and prior knowledge. However, 
to the best of our knowledge, all of them are based on purely supervised learning, which is notoriously weak under distribution shifts. IMU sensor data varies widely with different devices and users, sometimes the sensor data drifts over time. It is hard to control the distribution variability when the supervised algorithms are deployed in diverse applications. A rich and diverse database such as \textit{RoNIN}\cite{herath2020ronin} database can alleviate the problem to some extent, but it is cumbersome to collect such a big database and there are always scenarios that the database does not include and therefore the supervised model cannot capture their characteristics. 

In order to mitigate the challenge of distribution shift in real-world, we propose a geometric constraint, rotation-equivariance, that can improve generalizability of deep model in training phase and help the deep model to learn from shifted sensor data at inference time. Inspire by the Heading-Agnostic Coordinate Frame (HACF) presented in RoNIN\cite{herath2020ronin}, we define rotation-equivariance that when the IMU sequence in HACF is rotated around Z axis by a random angle, the corresponding ground-truth trajectory should be transformed by the same horizontal rotation. Under this assumption, we propose an auxiliary task that minimizing angle error between deep model prediction for rotated IMU data and rotated prediction of the original data. In experiments we validate that the auxiliary task improves model robustness in training phase when it is jointly optimized with the supervised velocity loss. During inference time, we formulate the auxiliary task as a self-supervised learning problem alone, that we update model parameters based on auxiliary losses generated by test samples to adapt the model to the distribution of given test data. This process is named as Test-Time Training (TTT)\cite{sun2020test}. Empirical results of TTT indicate the proposed self-supervision task brings substantial improvements at inference time. Furthermore, we introduce deep ensembles, a promising approach for simple and scalable predictive uncertainty estimation\cite{lakshminarayanan2016simple}. We show in experiments that the estimated uncertainty using deep ensembles is consistent with the error distribution. It helps us to develop adaptive TTT, that model parameters are updated when the uncertainty of prediction reaches a certain level. We compare different TTT strategies and study the relationship between update frequency and model precision.

In summary, our paper has the following three main contributions:

\begin{enumerate}[1)]
	\item We propose Rotation-equivariance-supervised Inertial Odometry (RIO) and demonstrate that rotation-equivariance can be formulated as an auxiliary task with powerful supervisory signal in training phase.
	\item We employ TTT based on rotation-equivariance for learning-based inertial odometry and validate that it helps to improve the generalizability of RIO.
	\item We introduce deep ensembles as a practical approach for uncertainty estimation, and utilize the uncertainty result as indicators for adaptively triggering TTT.
\end{enumerate}

The remainder structure of this paper is: we first give an overview on previous work regarding inertial odometry algorithms and related self-supervised tasks. Then we introduce our method and finally present experiments and evaluations.

	\section{Related work}

Roughly, there are three types of inertial odometry algorithms: i) double integration-based analytical solutions\cite{titterton2004strapdown, wu2005strapdown, bortz1971new}; ii) constrained model with additional assumptions\cite{park2010zero,foxlin2005pedestrian,ho2016step,jimenez2009comparison,solin2018inertial,hostettler2016imu} and iii) data-driven methods\cite{yan2018ridi,chen2018ionet,herath2020ronin,liu2020tlio,sun2021idol,wang2021pose}. 

Conventional strap-down inertial navigation system is to use double integration of IMU readings to compute positions\cite{titterton2004strapdown}. Many analytical solutions\cite{wu2005strapdown, bortz1971new} have been studied to promote the performance of the system. However, double integration leads to exploded cumulative error if there are signal biases. It requires high-precision sensors which are expensive and heavy, and typically are instrumented with aircrafts, automobiles and submarines. 

For consumer-grade IMUs that are small and cheap but have lower accuracy, a variety of constrained models with different assumptions emerged\cite{jimenez2009comparison} and mitigated error drifts to some extent. \cite{park2010zero,foxlin2005pedestrian} resort to shoe-mounted sensors to detect zero velocity for limiting velocity errors. \cite{ho2016step} proposes step-detection and step-length estimation algorithms to estimate walking distance under regular gait hypothesis. Inertial odometry models fused with available measurements by Extended Kalmann Filter (EKF) are presented in \cite{solin2018inertial,hostettler2016imu}.\cite{solin2018inertial} requires observations such as position fixes or loop-closures. \cite{hostettler2016imu} suppose negligible acceleration of the device equipmented with IMU.     

Data-driven methods further broaden applicable scenarios of IMUs and relax condition limitations. RIDI\cite{yan2018ridi} and PDRNet\cite{asraf2021pdrnet} proposes to estimate robust trajectories of natural human motions with supervised training in a hierarchical way. RIDI\cite{yan2018ridi} develop a cascaded regression model that first use a support vector machine to classify IMU placements and then type-specific support vector regression models to estimate velocities. PDRNet\cite{asraf2021pdrnet} employ a smartphone location recognition network to distinguish smartphone locations and then use different models trained for different locations for inference. IONet\cite{chen2018ionet} and RoNIN\cite{herath2020ronin} using unified deep neural networks provide more robust solutions that work in highly dynamic conditions. They show direct integration of estimated velocities helps with limiting error drifts and a unified deep neural network model is capable to generalize to various motions. TLIO\cite{liu2020tlio} introduces a stochastic cloning EKF coupled with the neural network to further reduce position drifts. IDOL\cite{sun2021idol} and \cite{wang2021pose} are recent deep learning-based works that release heavy dependent of device orientation. IDOL designs an explicit orientation estimation module relied on magnetometer readings and \cite{wang2021pose} propose a novel loss formulation to regress velocity from raw inertial measurements.  

Our work is in line with data-driven inertial odometry research that focuses on mitigating challenge of distribution shift in real-world. We propose rotation-equivariance as a self-supervision scheme to improve model generalizability and learn from unlabeled data. \cite{chen2019motiontransformer} proposes MotionTransformer framework that uses a shared encoder to transform inertial sequences into a domain-invariant hidden representation with generative adversarial networks. They focus on domain adaptation for long sensory sequences from different domains. Our method mainly deal with distribution shifts over one sensory sequence, and we show obvious improvements with the help of proposed self-supervision task. Notably, our work is a flexible module that can be combined with many other deep learning based approaches like RoNIN, TLIO and IDOL. 

Self-supervised tasks provide surrogate supervision signals for representation learning. Learning with self-supervision gains increased interest to improve model performance and avoid intensive manual labeling effort. Many vision tasks utilize self-supervision for pre-training\cite{Newell_2020_CVPR} or multitask learning\cite{ren2018cross}. \cite{Zhou_2017_CVPR} uses view synthesis as supervisor to learn depth and ego-motion from unstructured video. \cite{agrawal2015learning} shows that ego-motion-based supervision learns useful features for multiple vision problems. \cite{komodakis2018unsupervised} demonstrates that predicting image rotations is a promising self-supervised task for unsupervised representation learning. \cite{sun2020test} uses the image rotation task and creates self-supervised learning problem at test time. They validate their approach with object recognition and show substantial improvements under distribution shifts. 

	\section{RIO}

\subsection{Rotation-equivariance}
\label{sub_sec:rot_equ}
Our goal is to develop a self-supervised method to improve the robustness of inertial odometry, and make the model perform well in various scenarios.  We observe that the trajectory should be rotated in the same way as the IMU data in HACF when it is rotated around z-axis by a certain angle. We name this property as rotation-equivariance. It provides the benefit to learn a robust inertial odometry.

Specifically, for a sequence of accelerometer data in a world coordinate frame, namely acceleration $A = \{\vec a_t\}_{t=1}^n$ with $\vec a_t \in \mathbb{R}^3$, and gyroscope data for the same period in the same coordinate frame, angular velocity $\Omega = \{\vec \omega_t\}_{t=1}^n$ with $\vec \omega_t \in \mathbb{R}^3$, we randomly select a horizontal rotation $Rot(.|\phi)$ that rotates $A$ and $\Omega$ by $\phi$ degrees around $z$ axis, notate as $A^{\phi}$ and $\Omega^{\phi}$. The neural network model $F(\cdot)$ takes acceleration $A$ and angular velocity $\Omega$ as input and yields a velocity estimation $\vec v$ as output:
\begin{equation}
	\vec v = F(A, \Omega|\theta), 
	\label{eq:important}
\end{equation}
where $\theta$ are learnable parameters of model $F(\cdot)$. With rotation-equivariance, given velocity estimation $\vec v_1 = F(A, \Omega|\theta)$, $\vec v_2 = F(A^{\phi}, \Omega^{\phi}|\theta)$, there is only a horizontal rotation $\phi$ between $v_1$ and $v_2$. That is, if operator $Rot(.|\phi)$ is applied to velocity $v_1$ and then get the rotated velocity $v_1^{\phi}$, we expect $v_1^{\phi}=v_2$. Negative cosine similarity \cite{chen2021exploring} is employed to evaluate the difference between the two velocities:
\begin{equation}
	\mathcal{D}( v_1^{\phi}, v_2) = - \frac{<v_1^{\phi}, v_2>}{\lVert v_1^{\phi} \rVert_2 \cdot \lVert v_2 \rVert_2},
	\label{eq:cosine_sim}
\end{equation}
where $ <\cdot, \cdot> $ denotes the inner product between vectors. Therefore, we define a self-supervised auxiliary task, that given a set of $N$ training IMU samples $S = \{A_i, \Omega_i\}_{i=1}^N$, the neural network model must learn to solve the self-supervised training objective:
\begin{equation}
	\min_{\theta} \frac{1}{N}\sum_{i=1}^N\mathcal{L}(A_i, \Omega_i, \theta),
	\label{eq:general_loss}
\end{equation}
Where the loss function $\mathcal{L}(A_i, \Omega_i, \theta)$ is defined as:
\begin{equation}
	\frac{1}{K}\sum_{j=1}^K\mathcal{D}(F(A_i^{\phi_j}, \Omega_i^{\phi_j}|\theta), Rot(F(A_i, \Omega_i|\theta)|\phi).
	\label{eq:auxiliary_loss}
\end{equation}

\begin{figure}
	\centering
	\includegraphics[width=8cm]{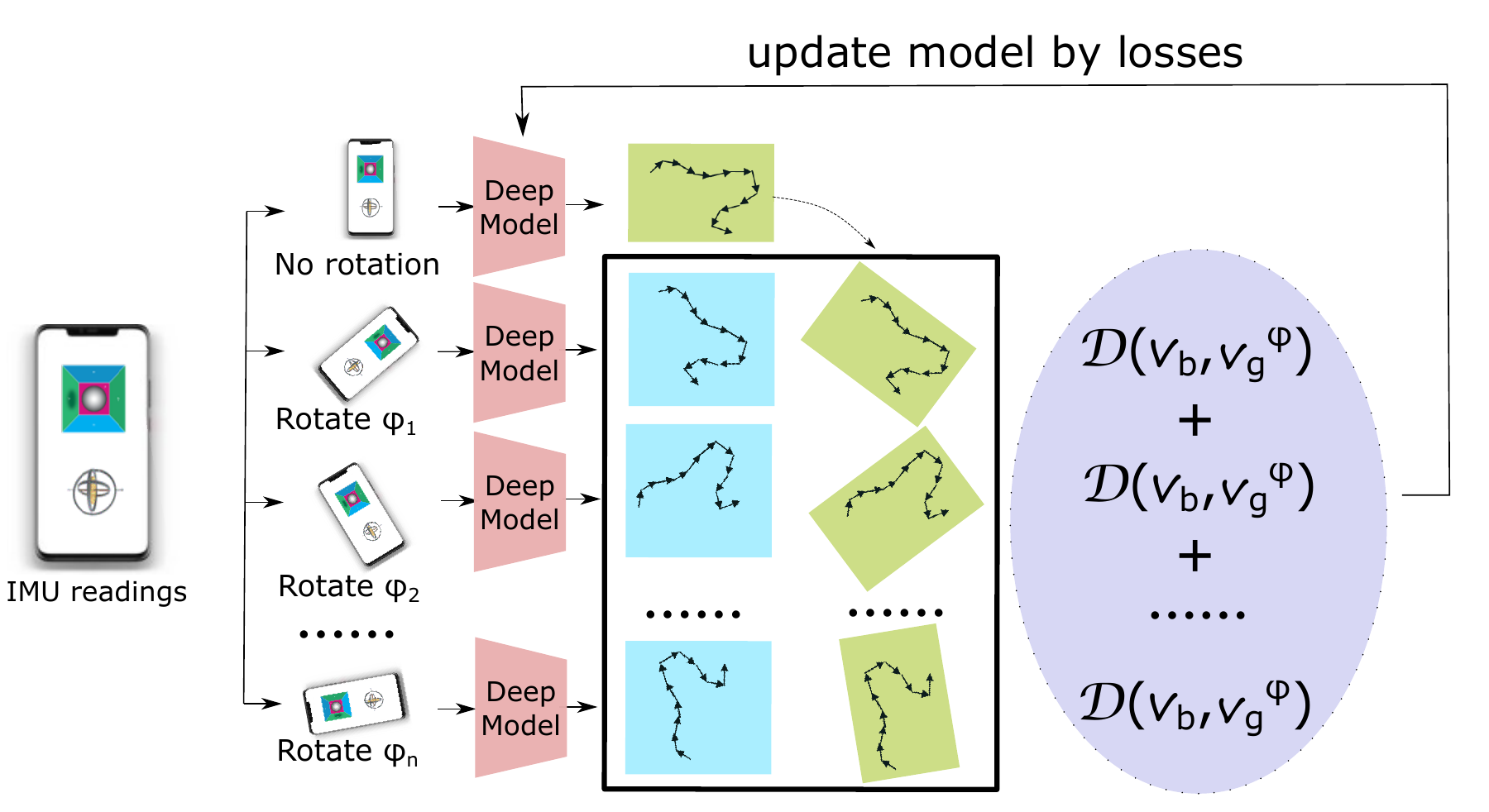}
	\vspace{-2ex}
	\caption{\textbf{Schematic illustration of proposed RIO.} For IMU readings, random angles are selected to generate rotated IMU data. The same deep model is applied to the original and rotated data to estimate trajectories. Estimated trajectory of the original data is rotated by the same set of angles. Estimated trajectories of rotated data are compared with corresponding rotated trajectory estimated by the original data to compute losses and update the deep model.}
	\label{fig:system}
\end{figure}

In the following subsections we describe how the self-supervised auxiliary task helps with model training and inference. 
\begin{algorithm}
	\caption{Joint-Training}
	\begin{algorithmic}[1]
		\For {$X, \vec v^{gt}$ in loader} \Comment{load a batch with n samples}
		\For {each $X_i, \vec v_i^{gt}$} {parallel}
		\State Random select a angle $\phi_i$
		\State $X_i^{\phi_i} = Rot(X_i|\phi_i)$ \Comment{compute conjugate input} 
		\State $\vec v_i=F(X_i|\theta)$
		\State $\vec v_i^c=F(X_i^{\phi_i}|\theta)$ \Comment{compute outputs}
		\State $l_{v_i} = \sum\limits_{x,y,z}(\vec v_i-\vec v_i^{gt})^2$ \Comment{compute velocity loss}
		\State $\vec v_i^{\phi_i} = Rot(\vec v_i|\phi_i)$ \Comment{prepare rotated output}
		\If {$\lVert v_i \rVert_2 >0.5$}
		\State $l_{ssl_i} = \mathcal{D}(v_i^{\phi_i}, v_i^c)$ \Comment{compute as \eqpref\eqref{eq:cosine_sim}}
		\Else
		\State $l_{ssl_i} = 0$ \Comment{set the loss to zero}
		\EndIf
		\EndFor
		\State $L = \sum\limits_il_{v_i} + \sum\limits_il_{ssl_i}$
		\State Update $\theta$ using Adam
		\EndFor
	\end{algorithmic} 
\end{algorithm}


\subsection{Joint-Training}
\label{sub_sec:joint}
In the training phase, we optimize the auxiliary loss (see \eqpref\eqref{eq:auxiliary_loss}) with velocity losses jointly. With the auxiliary task, the neural network model is encouraged to produce velocity estimations with a certain relative geometric relationship. However, it is unrealistic for the model to learn the magnitude and direction of the velocity in a consistent coordinate frame only with the auxiliary task. Jointly, we adopt the robust stride velocity loss to supervise the model. Stride velocity losses compute mean square error between the model output $\vec v_t$ at time frame $t$ and the ground truth velocity $\vec v_t^{gt}$ which is calculated by the average velocity over the sensor input time stride. In practice, we take one second sensor data as input and calculate the corresponding average velocity as supervisor.

To train the model on both tasks, we create conjugate data for each training input data and organize them as data pairs. IMU data is sampled at 200 Hz and we take every 200 continuous frames as input. In other words, we use IMU data last for 1 second as input. IMU data and ground truth trajectories are transformed into the same HACF as mentioned in RoNIN\cite{herath2020ronin}. And ground truth velocity is then calculated as the displacement of the corresponding time divide by the time length according to the ground truth trajectory in the HACF. More clearly, at frame $i$, input $X_i = (\{\vec a_t\}_{t=i-199}^i, \{\vec \omega_t\}_{t=i-199}^i) $, and ground truth velocity $\vec v^{gt}_i =\vec P_i –\vec P_{i-199}$ where $\vec P_i$ is the position on the trajectory. For each input $X_i$, select a random angle $\phi_i~(0<\phi_i\leqslant2\pi)$ to horizontally rotate accelerations and angular velocities in $X_i$ and get the conjugate data $X_i^{\phi_i}$. $X_i$ and its conjugate $X_i^{\phi_i}$ are processed by the neural network model $F$ and get two outputs as $\vec v_i = F(X_i|\theta)$ and $\vec v_i^c = F(X_i^{\phi_i}|\theta)$. For output $\vec v_i$, calculate the stride velocity loss as $(\vec v_i-\vec v_i^{gt})^2$. As shown in \secpref\ref{sub_sec:rot_equ}, rotate $\vec v_i$ around z axis by $\phi_i$ that $\vec v_i^{\phi_i} = Rot(\vec v_i|\phi_i)$ and calculate the negative cosine similarity between $\vec v_i^{\phi_i}$ and $\vec v_i^{c}$ as the loss for the self-supervised auxiliary task. To avoid ambiguous orientation of the velocity when stationary, we ignore the auxiliary loss when velocity magnitude is no more than $\mathrm{0.5~m/s}$. The pseudo-code of joint training can found in Algorithm 1.

\subsection{Adaptive TTT}
\label{sub_sec:ada_ttt}

At test time, we propose adaptive TTT based on rotation-equivariance and uncertainty estimation. It helps improve model performance on unseen data which has a large gap with training data. For test samples, we create conjugate data pairs the same as in training phase. With the self-supervised auxiliary task presented in \secpref\ref{sub_sec:rot_equ}, we calculate the auxiliary loss to update $\theta$ of neural network model $F(\cdot|\theta)$ before making predictions. For IMU data that arrive in an online stream, we adopt the online version that keep the state of the updated parameters for a while and restore the initial parameters in specific situations. 

While properly updated models can make substantial improvements under distribution shifts, their performance on original distribution may drop dramatically if TTT updates the parameters in an inappropriate way. The proposed auxiliary task cannot capture accurate losses when objects moving with an ambiguous orientation like moving slowly or stationary. At inference time, the velocity threshold used in training phase is not enough to ensure stable and reliable updates since batch data to optimize model is from a continuous period of time and they tend to have ambiguous orientation at the same time. Therefore, we introduce uncertainty estimation to assist with determining the right time to update or restore model parameters.

\textbf{Uncertainty estimation} We use deep ensembles to provide predictive uncertainty estimations that are able to express higher uncertainty on out-of-distribution examples\cite{lakshminarayanan2016simple}. We adopt a randomization-based approach, that with random initialization of the neural network models parameters and random shuffling of the training data to get individual ensemble models. 

Formally, we randomly initialize $M$ neural network models $\{F(X|\theta_m\}_{m=1}^{M}$with different parameters $\theta_m $ that each of them parameterize a different distribution on the outputs. Each model converges through an independent optimization path with training data random shuffling. For convenience, assume the ensemble is a Gaussian distribution and each model prediction $p_{\theta_m} = F(X|\theta_m)$ represents a sample from the distribution. We approximate the prediction uncertainty as the variance of sampled predictions that $\sigma^2=\frac{1}{M}\sum\limits_m(p_{\theta_m}-p_*)^2$ where $p_* = \frac{1}{M}\sum\limits_m p_{\theta_m}$. 

For our inertial odometry model, we get velocity $\vec v_{\theta_m}$ from model $F(X|\theta_m)$, and the velocity variance can be calculated with corresponding sampled estimations. We show in experiments that velocity variance based on deep ensembles well indicates models confidence level for the estimation. The velocity variance is used as prediction uncertainty indicators to determine when to update or restore model parameters.

\textbf{TTT strategy} Further, we propose an adaptive TTT strategy based on uncertainty estimations. First, we stop updating model parameters when velocity estimations have a high confidence level. When objects move with ambiguous orientation, the auxiliary loss tends to be large, however, velocity variance is not necessary to be large and tends to be small. It avoids overhead updating and only updates models when necessary. 

Second, we need to know when to reset models. Models will drift a lot if there is inappropriate updating. We hope to keep the state of updated parameters if the motion is continuous. However, if the motion switches to a different mode, IMU data distribution will change a lot and the updated model performance may be worse than the original model on the unseen data. Meanwhile, from a simple observation that in most cases there is a stationary or nearly stationary zone between two different motion modes, we propose to restore original model parameters when objects stationary or nearly stationary. We use velocity uncertainty to capture these moments in that the inertial odometry model tends to have an absolute high confidence level when stationary or nearly stationary.

To do inference at test time, the neural network model is first initialized with pre-trained parameters $\theta^*$. Test samples that with 200 frames IMU data are sampled every 10 frames at 20 Hz, and when 128 test samples arrive, we make them in a batch $X$ for test-time training. For convenience, we select four degrees $\{72^{\circ}, 144^{\circ}, 216^{\circ}, 288^{\circ}\}$ to create conjugate inputs the same way as in the training phase. With the original and conjugate inputs, we can get velocity estimations from the model. Denoting original outputs as $V$ and conjugate outputs as $V^c$, rotate original outputs by corresponding angle $\phi$ and get $V^{\phi}$. The auxiliary loss $\mathcal{L}$ is calculated the same as in the training phase:

\begin{equation}
	\mathcal{L}(v_j^{\phi_i}, v_j^{c_i})=
	\begin{cases}
		\mathcal{D}(v_j^{\phi_i}, v_j^{c_i})& \text{$\lVert v_j \rVert_2 >0.5$} \\
		0& \text{$\lVert v_j \rVert_2 \leq 0.5$}
	\end{cases}
	\label{eq:important}
\end{equation}

For every batch of data, we update the model at most 5 times. With deep ensemble-based uncertainty estimation, velocity uncertainty is estimated as the outputs variance of three independent pre-trained models, denoted as $\sigma_v^2$. According to the adaptive TTT policy, we stop updating the model if the average velocity variance $\overline{\sigma_v^2}$ is smaller than a certain value; restore original parameters if the minimal velocity variance $min(\sigma_{v_i}^2) $ is absolute small. In practice, we stop updating if $\overline{\sigma_v^2} < 0.04$ and restore parameters if any $min(\sigma_{v_i}^2) < 1e-4$. The pseudo-code of adaptive TTT can be found in the supplementary.

	\section{Evaluations}

We evaluate our proposed method in this section. Our main purpose is to verify that the proposed auxiliary task based on rotation-equivariance helps to improve models robustness and accuracy. In order to eliminate the influence of model architecture and datasets used for training, we adopt a consistent mature architecture and datasets for all models used for evaluation. All models are with ResNet-18 \cite{he2016deep} backbone and we use the largest smartphone-based inertial navigation database provided by RoNIN to train models\cite{herath2020ronin}. With different supervision tasks in training phase and different strategies at inference time, we demonstrate that the proposed auxiliary task helps the model outperform the existing state-of-the-art method.

\begin{table*}[t]
	\centering
	\begin{tabular}{ccccccc}
		\toprule
		Database & Metric & R-ResNet & B-ResNet & J-ResNet  & B-ResNet-TTT & J-ResNet-TTT \\
		\midrule
		\multirow{3}*{RoNIN}& ATE ($m$) & 5.14 & 5.57 & \textbf{\textcolor{red}{5.02}} & \textbf{{5.05}} & \textbf{5.07} \\
		& RTE ($m$) & 4.37 & 4.38 & \textbf{4.23} & \textbf{\textcolor{red}{4.14}} & \textbf{{4.17}} \\
		& D-drift & 11.54$\%$  & 9.79$\%$ & \textbf{9.59$\%$} & \textbf{\textcolor{red}{8.49$\%$}} & \textbf{9.10$\%$} \\
		\midrule
		\multirow{3}*{OXIOD}& ATE ($m$) & 3.46 & 3.52 & 3.59 & \textbf{\textcolor{red}{2.92}} & \textbf{{2.96}} \\
		& RTE ($m$) & 4.39 & 4.42 & 4.43 & \textbf{\textcolor{red}{3.67}} & \textbf{{3.74}} \\
		& D-drift & 20.67$\%$ & 19.68$\%$ & \textbf{17.43$\%$} & \textbf{\textcolor{red}{15.50$\%$}} & \textbf{{15.98$\%$}} \\
		\midrule
		\multirow{3}*{RIDI}& ATE ($m$) & 1.33 & 1.19 & \textbf{1.13} & \textbf{1.04} & \textbf{\textcolor{red}{1.03}} \\
		& RTE ($m$) & 2.01 & 1.75 & \textbf{1.65} & \textbf{1.53} & \textbf{\textcolor{red}{1.51}} \\
		& D-drift & 10.50$\%$ & 7.99$\%$ & \textbf{7.61$\%$} & \textbf{\textcolor{red}{6.89$\%$}} & \textbf{{6.93$\%$}} \\
		\midrule
		\multirow{3}*{IPS}& ATE ($m$) & 1.60 & 1.84 & \textbf{1.67} & \textbf{\textcolor{red}{1.55}} & \textbf{\textcolor{red}{1.55}} \\
		& RTE ($m$) & 1.52 & 1.68 & \textbf{1.65} & \textbf{\textcolor{red}{1.46}} & \textbf{{1.47}} \\
		& D-drift & 8.38$\%$ & 7.66$\%$ & 7.96$\%$ & \textbf{\textcolor{red}{5.93$\%$}} & \textbf{6.75$\%$} \\
		\bottomrule
	\end{tabular}
	\vspace{-2ex}
	\caption{\textbf{Performance evaluation.} We compare five methods: R-ResNet, B-ResNet, J-ResNet with standard inference pipeline; B-ResNet and J-ResNet with TTT. Methods are evaluated on the test data of four datasets: RoNIN, OXIOD, RIDI, and IPS. Best results are highlighted in red per row.}
	\label{tab:main_result}
\end{table*}

\begin{figure}
	\centering
	\includegraphics[width=8.8cm]{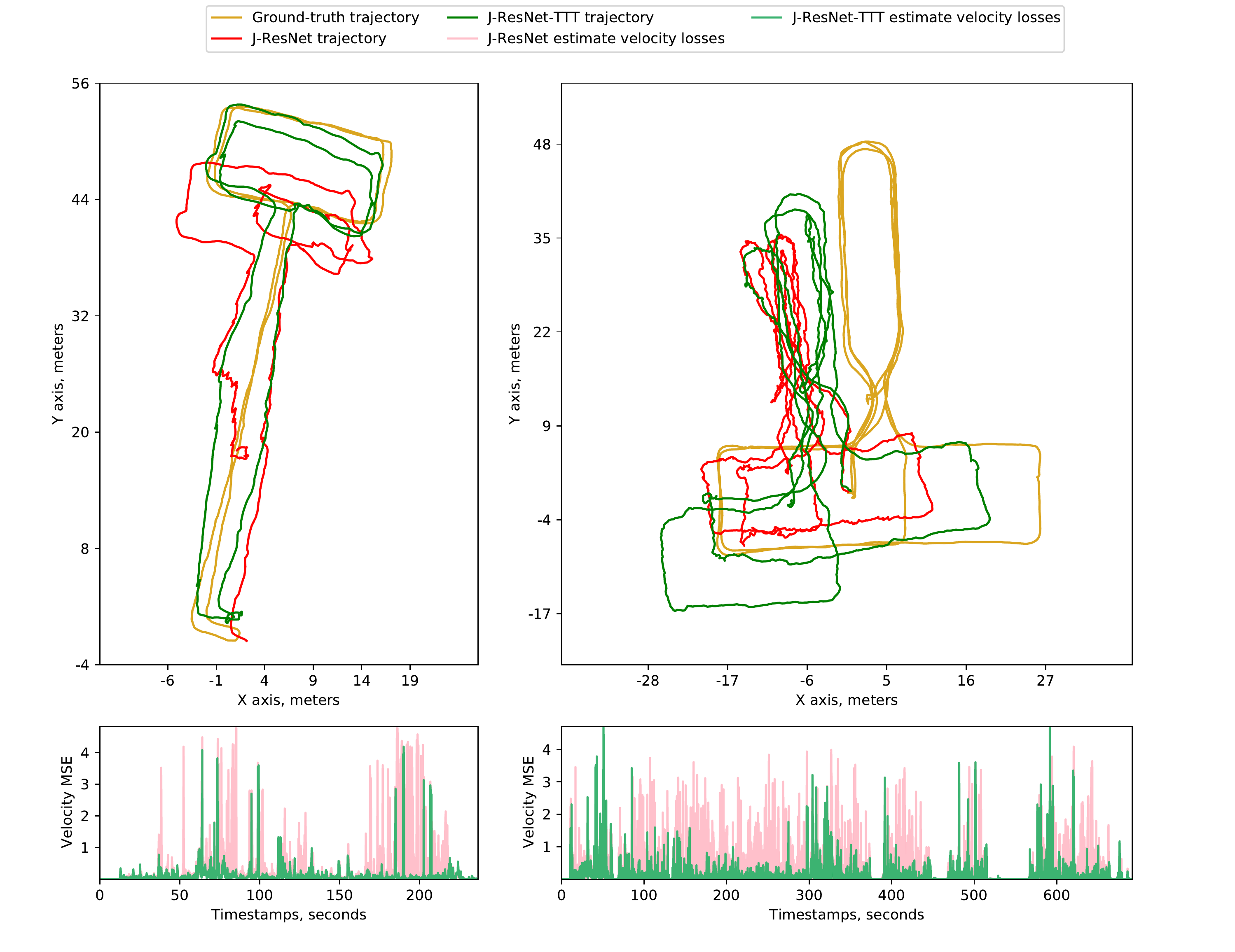}
	\vspace{-6ex}
	\caption{Comparison of example trajectories. The left is a trajectory from \textit{IPS} database and the right is from \textit{RoNIN}. For both, TTT strategy based on rotation-equivariance descreased velocity MSE and resulted in better trajectory estimations. More examples are in the appendix.}
	\label{fig:example_traj}
\end{figure}


\textbf{Network details} We adopt Resnet-18 backbone since ResNet-18 model achieved the highest accuracy on multiple datasets shown by RoNIN. We replace Batch Normalization (BN) with Group Normalization (GN) in that the trained model is going to be used in TTT where training with small batches. BN that uses estimated batch statistics has been shown to be ineffective with small batches whose statistics are not accurate. GN that uses channels group statistics is not influenced by batch size and results in similar results as BN on inertial odometry problem. As we propose in \cref{sub_sec:joint}, we train a model denoted as \textit{J-ResNet} with joint-training setting following Algorithm 1. 

We use the RoNIN model with ResNet-18 backbone as a baseline. While RoNIN publish a pre-trained ResNet model, denoted as \textit{R-ResNet}, which is exactly the one they claimed in \cite{herath2020ronin}, it is a model using BN and trained with the whole RoNIN dataset. They only publish half of the whole database due to privacy limitation. For fair comparison, we re-train a model using GN with the public database as a baseline. Other implementations are exactly the same as they claimed in \cite{herath2020ronin}. We denote the re-trained model as \textit{B-ResNet}.

\textbf{Databases} Models are evaluated with three popular public databases for inertial odometry: \textit{OXIOD}\cite{chen2018oxiod}, \textit{RoNIN}\cite{herath2020ronin} and \textit{RIDI}\cite{yan2018ridi}, and one database collected in different scenarios by ourselves, \textit{IPS}. Collecting details are presented in the supplementary. For trajectory sequences in \textit{OXIOD} \textit{RIDI} and \textit{IPS}, the whole estimated trajectory is aligned to the ground-truth trajectory with Umeyama algorithm \cite{umeyama1991least} before evaluation. For \textit{RoNIN} whose sensor data and ground-truth trajectory data are well calibrated to the same global frame, we directly use the reconstructed trajectory to compare with the ground-truth.

We evaluate neural networks \textit{J-ResNet} and \textit{B-ResNet} with two different approaches. One is the standard neural network inference pipeline which is the same as in IONet\cite{chen2018ionet}, RoNIN\cite{herath2020ronin}, and another use the adaptive TTT proposed in \cref{sub_sec:ada_ttt}. \textit{R-ResNet} is evaluated only with standard pipeline since it is with BN as normalization layers and it cannot be optimized with small data batch.

\begin{figure*}
	\centering
	\includegraphics[width=18cm]{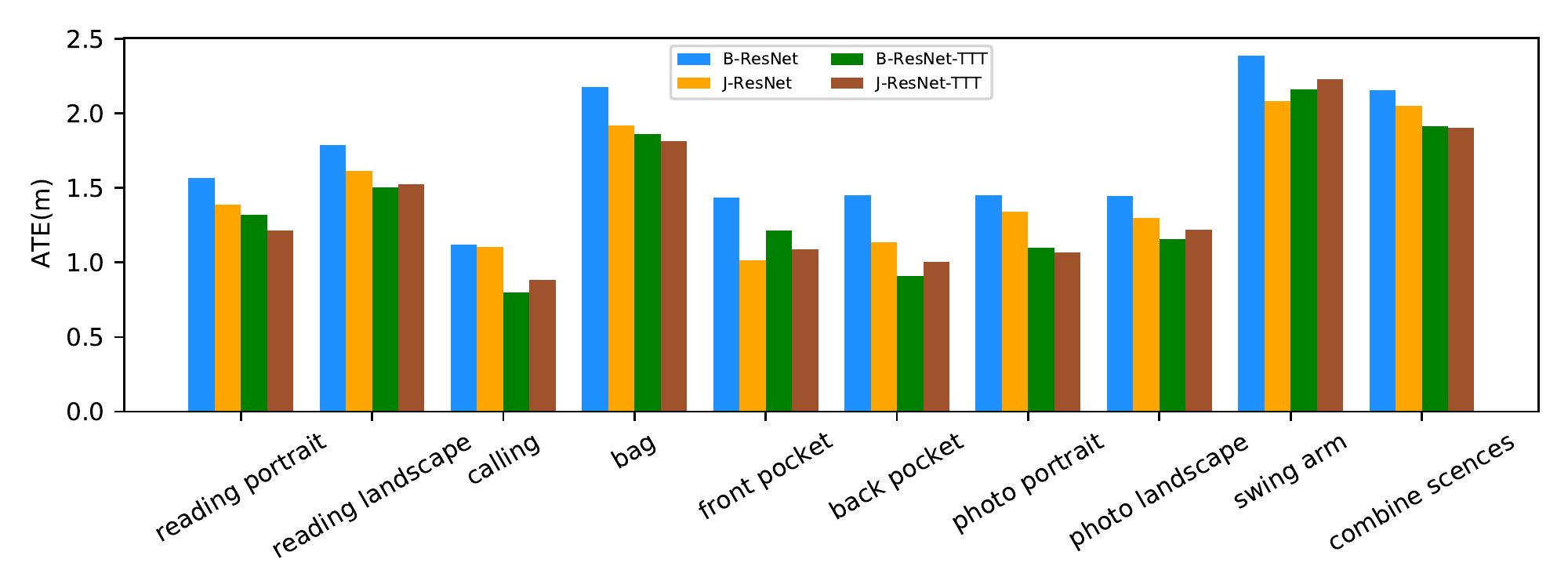}
	\vspace{-7ex}
	\caption{Comparison of performance on different scenarios included in \textit{IPS} database. In most scenarios, models with TTT strategy have a better performance and \textit{J-ResNet} is better than \textit{B-ResNet}.}
	\label{fig:scenarios}
\end{figure*}

\subsection{Metrics definitions}

 Three metrics are used for quantitative trajectory evaluation of inertial odometry model: Absolute Trajectory Error (ATE) and Relative Trajectory Error (RTE), and Distance drift (D-drift). ATE and RTE are standard metrics proposed in \cite{sturm2012benchmark}.

\textbf{\textit{ATE}} ($m$), is calculated as the average Root Mean Squared Error (RMSE) between the estimated and ground-truth trajectories as a whole. 

\textbf{\textit{RTE}} ($m$), is calculated as the  average RMSE between the estimated and ground-truth over a fixed length or time interval. Here we use time-based RTE the same as in RoNIN that we evaluate RTE over 1 minute.

\textbf{\textit{D-drift}}, is calculated as absolute difference between the estimated and ground-truth trajectory length divided by the length of ground-truth trajectory.

\subsection{Performance}

\cref{tab:main_result} is our main results. All subjects used to evaluate models do not present in training sets. Our evaluation of the \textit{R-ResNet} for \textit{RoNIN} test datasets is consistent with the report of \textit{RoNIN unseen sets} in \cite{herath2020ronin}. Other three datasets are not used in the training phase. \textit{R-ResNet} is trained with full \textit{RoNIN} dataset and we only use half of it which is published to train \textit{B-ResNet} and \textit{J-ResNet}. Therefore, we evaluate \textit{R-ResNet} performance just for reference. \textit{B-ResNet} is a fair baseline and we compare other methods with it. 

The results show that \textit{J-ResNet} outperforms \textit{B-ResNet} on most databases. \textit{J-ResNet} reduces ATE by $9.96\%$, $4.47\%$ and $9.13\%$ on \textit{RoNIN}, \textit{RIDI} and \textit{IPS} databases, respectively. Notably, for \textit{RoNIN} database, \textit{J-ResNet} outperforms \textit{R-ResNet} which is trained with twice as much training data.

\textit{B-ResNet-TTT} outperforms \textit{B-ResNet} on all databases by a significant margin. It reduces ATE by $9.29\%$, $17.04\%$, $11.84\%$ and $15.59\%$ on \textit{RoNIN}, \textit{RIDI}, \textit{OXIOD} and \textit{IPS} databases, respectively. For \textit{J-ResNet-TTT}, it reduces ATE by $17.55\%$, $9.43\%$ and $7.06\%$ on \textit{OXIOD}, \textit{RIDI} and \textit{IPS}, and it has a comparable performance on \textit{RoNIN} comparing to \textit{J-ResNet}. In a word, the adaptive TTT strategy proposed in \cref{sub_sec:ada_ttt} can further improve performance of \textit{B-ResNet} and \textit{J-ResNet}. 

Both models are trained with \textit{RoNIN} training database in training phase. \textit{J-ResNet} is trained with the auxiliary task and it already helps to improve performance on \textit{RoNIN} test database. We assume the auxiliary task is optimized in training phase for \textit{RoNIN} database that it does not significant improve model performance further with test-time training. For \textit{OXIOD}, \textit{RIDI} and \textit{IPS} which are novel databases for both models, adaptive TTT further improve models performance on all metrics. 

\cref{fig:example_traj} shows selected trajectories performance visualization of {\textit{J-ResNet} and {\textit{J-ResNet-TTT}. It shows estimate trajectories against the ground-truth of both models along with velocity estimation losses comparison. Velocity estimation losses are reduced a lot when there are large velocity losses of original model. It demonstrates that the auxiliary task used in adaptive TTT can help to optimize model at pivotal steps and result in a better trajectory estimation.

\subsection{Performance on Multiple Scenarios}

The proposed rotation-equivariance contributes differently in different scenarios. Although \textit{RoNIN} training database is the largest public inertial odometry database with rich diversity, model performance under certain scenarios can be improved by a large margin using RIO.

We compare model performance by scenarios in \textit{IPS} database and present results in \cref{fig:scenarios}. In different scenarios, devices with IMU sensors are mounted at different placements, and are handled with different ways. \cref{fig:scenarios} shows that in all scenarios, \textit{J-ResNet} outperforms \textit{B-ResNet}, and TTT version of both models improve even further. Under \textit{calling}, \textit{back pocket}, \textit{photo portrait} and \textit{photo landscape} scenarios, ATE of \textit{B-ResNet} can be reduced by $28.55\%$, $37.56\%$, $24.47\%$ and $19.92\%$, and for \textit{J-ResNet}, TTT version reduces ATE by $20.06\%$, $11.36\%$, $20.13\%$ and $6.02\%$, respectively. These four scenarios are not very common in daily life and may not show up as frequent as other poses in \textit{RoNIN} training database. Therefore, large improvements under unusual scenarios elucidate that adaptive TTT helps trained models to learn from novel data distribution and improve their performance under distribution shifts.

\subsection{TTT Strategy Analysis}

In this section, we evaluate the TTT strategy in isolation and explain why uncertainty estimations help with model performance improvement.

\begin{figure}
	\centering
	\includegraphics[width=8.5cm,height=6cm]{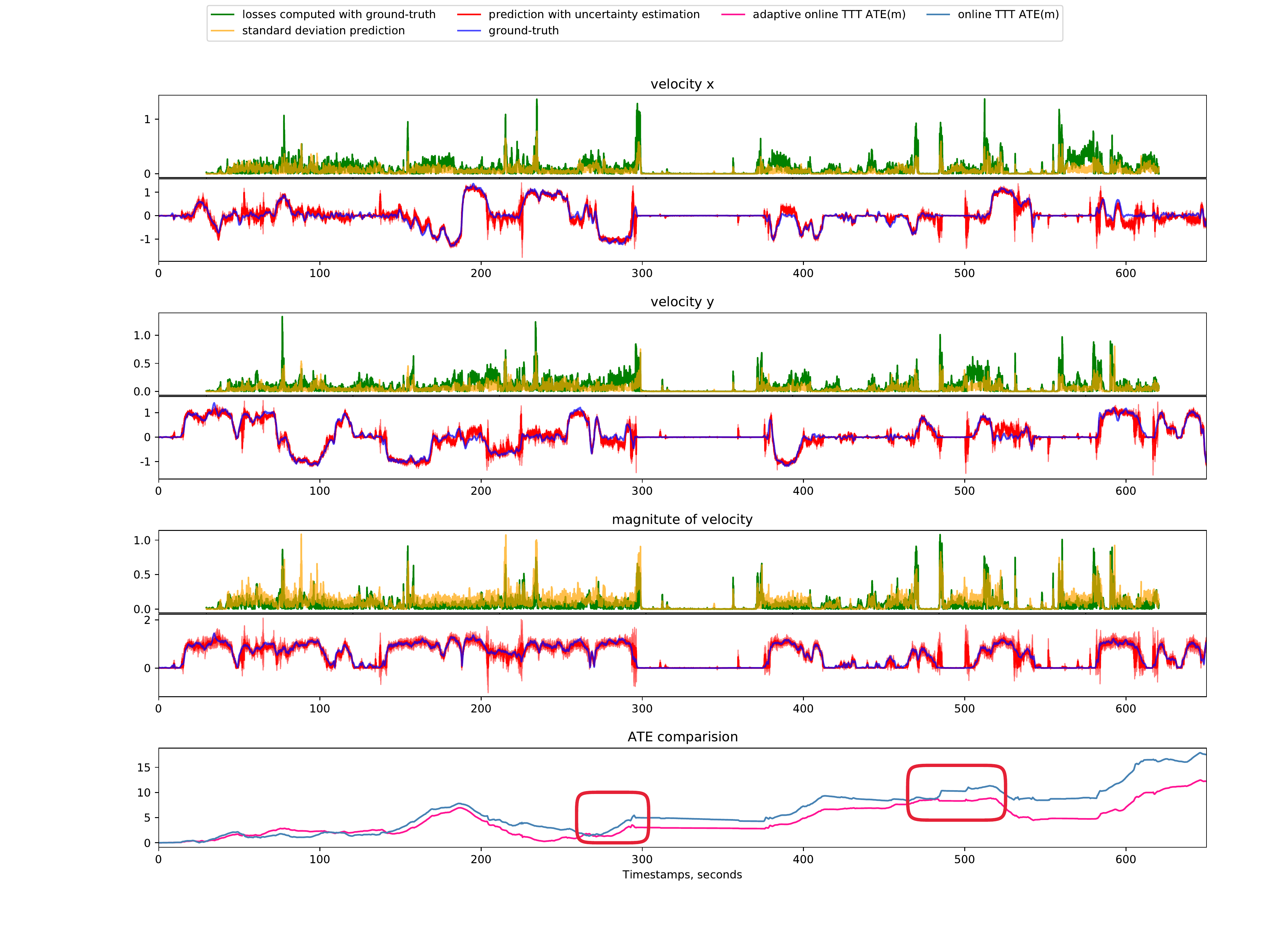}
	\vspace{-6ex}
	\caption{Example of the uncertainty estimation. Estimated velocity uncertainty well follows estimation errors. Adaptive TTT using uncertainty estimations leads to a lower ATE as the model is updated. More examples are in the appendix.}
	\label{fig:velocity_uncertainty}
\end{figure}

\begin{table}[t]
	\centering
	\begin{tabular}{ccccccc}
		\toprule
		Database & Metric & A-TTT & N-TTT \\
		\midrule
		\multirow{3}*{RoNIN}& ATE ($m$) & {{5.05}} &  \textcolor{red}{4.94} \\
		& RTE ($m$) & {\textcolor{red}{4.14}} & {{4.27}} \\
		& D-drift & {\textcolor{red}{8.49$\%$}} & {9.43$\%$} \\
		\midrule
		\multirow{3}*{OXIOD}& ATE ($m$) & {\textcolor{red}{2.92}} & {{3.50}} \\
		& RTE ($m$) & {\textcolor{red}{3.67}} & {{4.39}} \\
		& D-drift & {\textcolor{red}{15.50$\%$}} & {{19.55$\%$}} \\
		\midrule
		\multirow{3}*{RIDI}& ATE ($m$) &  \textcolor{red}{1.04} & {1.11} \\
		& RTE ($m$) &  \textcolor{red}{1.53} & {1.64} \\
		& D-drift & {\textcolor{red}{6.89$\%$}} & {{7.56$\%$}} \\
		\midrule
		\multirow{3}*{IPS}& ATE ($m$) & {\textcolor{red}{1.55}} & {1.63}\\
		& RTE ($m$) & {\textcolor{red}{1.46}} & {1.54} \\
		& D-drift & {\textcolor{red}{5.93$\%$}} &{6.73$\%$} \\
		\bottomrule
	\end{tabular}
	\vspace{-2ex}
	\caption{\textbf{TTT strategies comparision.} A-TTT has an obvious advantage over N-TTT for all metrics on four databases. }
	\label{tab:ttt_compare}
\end{table}

\textit{1) Is uncertainty estimation with deep ensemble reliable?}  \cref{fig:velocity_uncertainty} shows one trajectory velocity estimations against its ground-truth together with their uncertainty estimations and velocity estimation losses. It demonstrates that our predictive uncertainty well follows the estimation losses. As we expected, the predictive uncertainty decreases when velocity estimation losses decrease. With the uncertainty estimation, we can stop updating when it is below a certain level since the model is pretty accurate at this time. Notably, the predictive uncertainty decreases to zero when the magnitude of velocity is around zero. As mentioned in \cref{sub_sec:ada_ttt}, detection of stationary or nearly stationary zones is important for adaptive TTT in that it is the right time to restore original model parameters. 
 
\textit{2) Comparing adaptive TTT with others:}  Last row of \cref{fig:velocity_uncertainty} compares Adaptive TTT (A-TTT) ATE change over time with Naive TTT (N-TTT). N-TTT refers to the process that always update models according to losses of self-supervised task and ignore velocity uncertainty estimations. It keeps the latest updated model and does not restore original parameters over one continuous trajectory. \cref{fig:velocity_uncertainty} shows that ATE of N-TTT increase faster than A-TTT. There are two obvious time windows that ATE increase steeper with N-TTT, and during these time velocity is decreasing to zeros which means object is going to be stationary. With adaptive strategy, the model will restore original parameters and ATE will be suppressed. \cref{tab:ttt_compare} compares A-TTT and N-TTT on four databases and shows that the performance of A-TTT has obvious advantages over N-TTT.
									
	\section{Ablation Studies}

We conducted additional experiments with joint-training and TTT settings under ablation considerations.

\textbf{Models performance v.s. size of training data} We trained models in joint-training with different size of training datasets. Denote the neural network which is provided and published by RoNIN\cite{herath2020ronin} as \textit{100\% B-ResNet} since it is trained with the whole \textit{RoNIN} database. We trained models with 50\%, 30\% and 10\% data of the whole database in two ways as mentioned before, and evaluate their performance under different settings. \cref{fig:datasize} shows the comparison of different models. While \textit{B-ResNet} and \textit{J-ResNet} performances drop a lot as the training database becomes smaller, \textit{J-ResNet-TTT} with 30\% training database is still comparable to \textit{100\% B-ResNet}. However, \textit{J-ResNet-TTT} performance also drops a lot when using 10\% training databases.

\begin{figure}
	\centering
	\includegraphics[width=8cm]{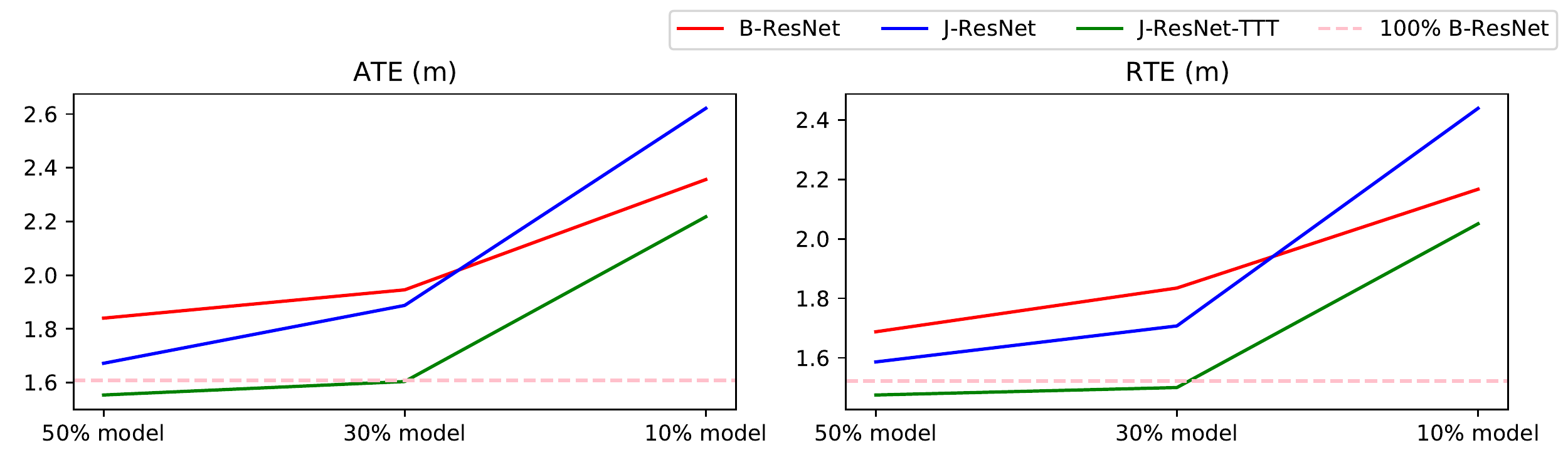}
	\vspace{-3ex}
	\caption{Impact of the \textbf{size of training data} on ATE and RTE. Methods are evaluated on \textit{IPS} database and compare them with the performance of \textit{100\% B-ResNet}.}
	\label{fig:datasize}
\end{figure}

\begin{figure}
	\centering
	\includegraphics[width=8cm]{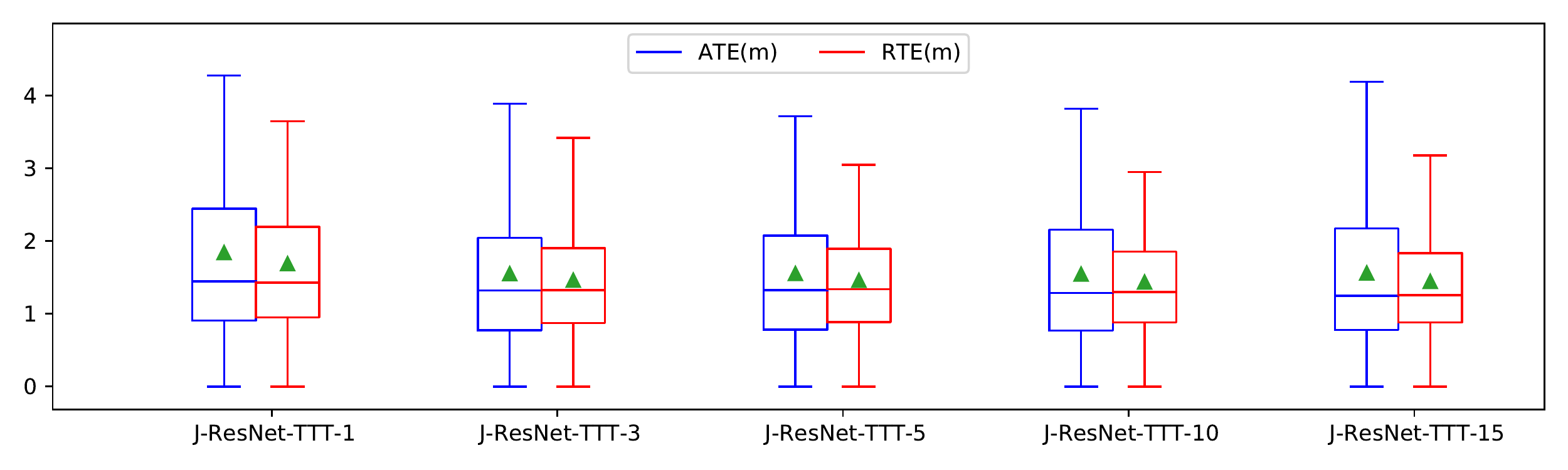}
	\vspace{-3ex}
	\caption{Impact of the \textbf{number of TTT iterations} on ATE and RTE. Methods are evaluated on \textit{IPS} database. }
	\label{fig:iterations}
\end{figure}

\textbf{Influence of updating iterations}  At test-time, model can be updated multiple times with one batch of data. \cref{fig:iterations} shows results of one model with different updating iterations from 1 to 15. There are obvious improvements when increase iterations from 1 to 5. However, more than 5 updates do not show obvious advantages and the model performance even degrade a little when updating 15 times one batch. More iterations cost more time and computing resource. Therefore, we recommend no more than 5 updates one batch during TTT.

	\section{Discussion and Conclusion}
In this paper we present a Rotation-equivariance-supervised Inertial Odometry (RIO) in order to improve performance and robustness of inertial odometry. The rotation-equivariance can be formulated as a self-supervised auxiliary task and can be applied both in training phase and inference stage. Extensive experiments results demonstrate that the rotation-equivariance task helps with advancing model performance under joint-training setting and will further improve model with Test-Time Training (TTT) strategy. Not only rotation-equivariance, there may be more equivariance (e.g., time reversal, mask auto-encoder of time series) that can be formulated as self-supervised task for inertial odometry. We hope our observation will enlighten future work in the aspect of self-supervise learning of inertial odometry.   

Further, we propose to employ deep ensemble to estimate the uncertainty of RIO. With uncertainty estimation, we develop adaptive TTT for evolving RIO at inference time. It thus can largely improve the generalizability of RIO. Adaptive TTT using the auxiliary task makes a model trained with less than one-third of the data outperforms the state-of-the-art deep inertial odometry model, especially under scenarios that the model does not see during the training phase. Adaptive online model update with uncertainty estimation is a practical way to improve deep model performance in real life applications. Uncertainty estimation based on deep ensemble gives reliable judgment on the output of deep models. Adaptive TTT can be implemented in different ways, either conservative or aggressive, for updating the model depending on the application scenarios.



	{\small
		\bibliographystyle{ieee_fullname}
		\bibliography{./reference/3d_pdr_ref}
	}
	\section*{Supplementary Materials}

\textbf{Models performance v.s. size of training data} We trained models in joint-training setting with different size of training datasets. Denote the neural network which is provided and published by RONIN as 100\% B-ResNet since it is trained with the whole RONIN database. We trained models with 50\%, 30\% and 10\% data of the whole database in two ways as mentioned before. And evaluate their performance under different settings. \cref{fig:datasize} shows the comparison of different models. While \textit{B-ResNet} and \textit{J-ResNet} performances drop a lot as the training database becomes smaller, \textit{J-ResNet-TTT} with 30\% training database is still comparable to 100\% B-ResNet. However, \textit{J-ResNet-TTT} performance also drops a lot when using 10\% training databases.

\textbf{Influence of updating iterations}  At test-time, model can be update multiple times with one batch of data. \cref{fig:iterations} shows results of one model with different updating iterations from 1 to 15. There are obvious improvements when increase iterations from 1 to 5. However, more than 5 updates do not show obvious advantages and the model performance even degrade a little when updating 15 times one batch. More iterations cause more time and computing resource consumption. Therefore, we recommend no more than 5 updates one batch during TTT.

At test-time, model can be update multiple times with one batch of data. \cref{fig:iterations} shows results of one model with different updating iterations from 1 to 15. There are obvious improvements when increase iterations from 1 to 5. However, more than 5 updates do not show obvious advantages and the model performance even degrade a little when updating 15 times one batch. More iterations cause more time and computing resource consumption. Therefore, we recommend no more than 5 updates one batch during TTT.

\textbf{Influence of updating iterations}  At test-time, model can be update multiple times with one batch of data. \cref{fig:iterations} shows results of one model with different updating iterations from 1 to 15. There are obvious improvements when increase iterations from 1 to 5. However, more than 5 updates do not show obvious advantages and the model performance even degrade a little when updating 15 times one batch. More iterations cause more time and computing resource consumption. Therefore, we recommend no more than 5 updates one batch during TTT.

At test-time, model can be update multiple times with one batch of data. \cref{fig:iterations} shows results of one model with different updating iterations from 1 to 15. There are obvious improvements when increase iterations from 1 to 5. However, more than 5 updates do not show obvious advantages and the model performance even degrade a little when updating 15 times one batch. More iterations cause more time and computing resource consumption. Therefore, we recommend no more than 5 updates one batch during TTT.

\begin{figure}
	\centering
	\includegraphics[width=8cm]{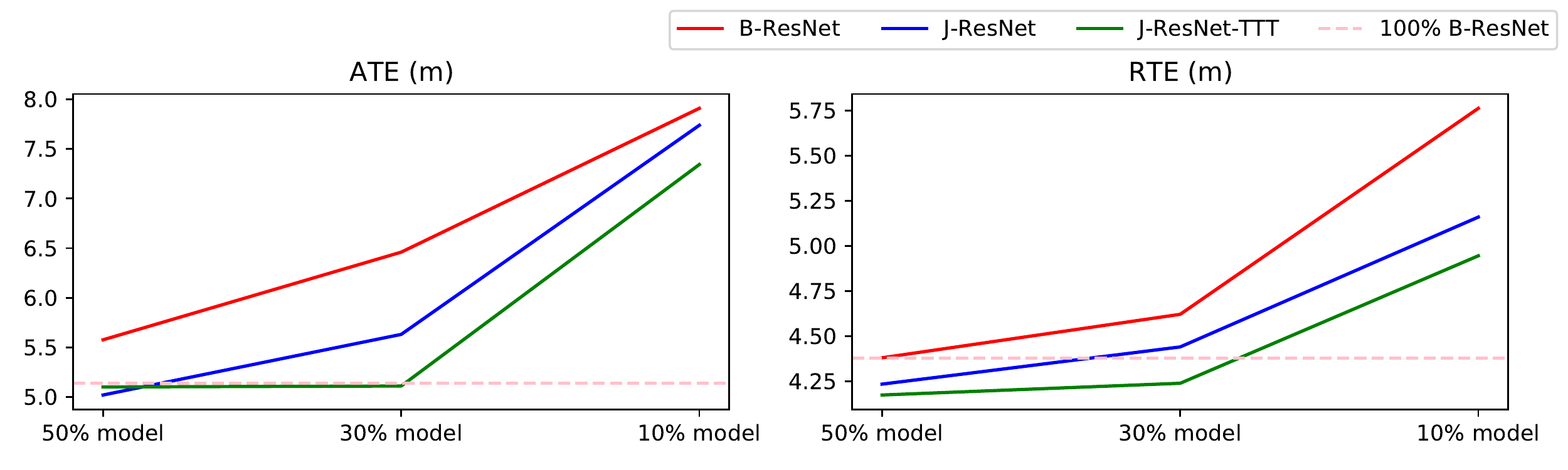}
	\vspace{-1ex}
	\caption{Impact of the number of TTT iterations on ATE and RTE}
	\label{fig:met_ronin}
\end{figure}

\begin{figure}
	\centering
	\includegraphics[width=8cm]{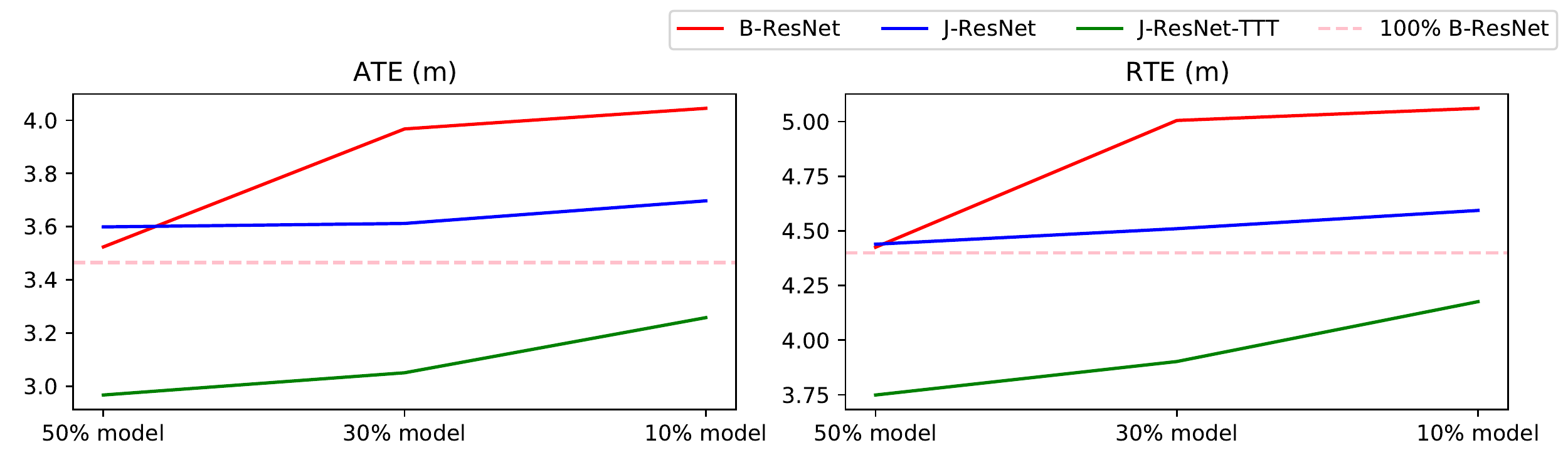}
	\vspace{-1ex}
	\caption{Impact of the number of TTT iterations on ATE and RTE}
	\label{fig:met_oxiod}
\end{figure}

\begin{figure}
	\centering
	\includegraphics[width=8cm]{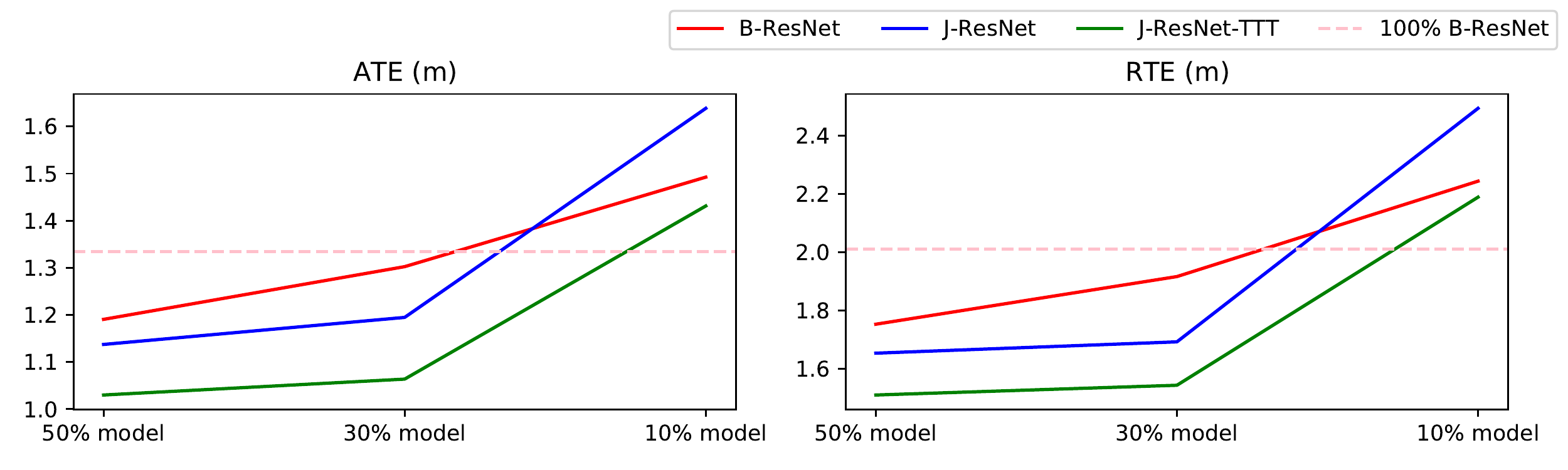}
	\vspace{-1ex}
	\caption{Impact of the number of TTT iterations on ATE and RTE}
	\label{fig:met_ridi}
\end{figure}

\begin{figure}
	\centering
	\includegraphics[width=8cm]{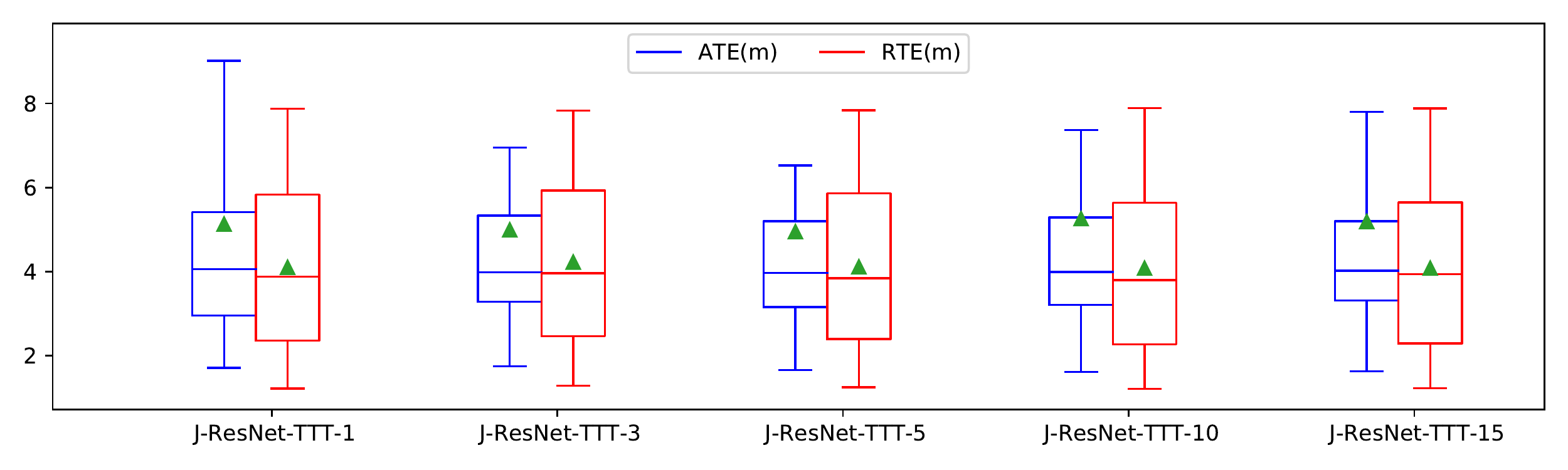}
	\vspace{-1ex}
	\caption{Impact of the number of TTT iterations on ATE and RTE}
	\label{fig:iter_ronin}
\end{figure}

\begin{figure}
	\centering
	\includegraphics[width=8cm]{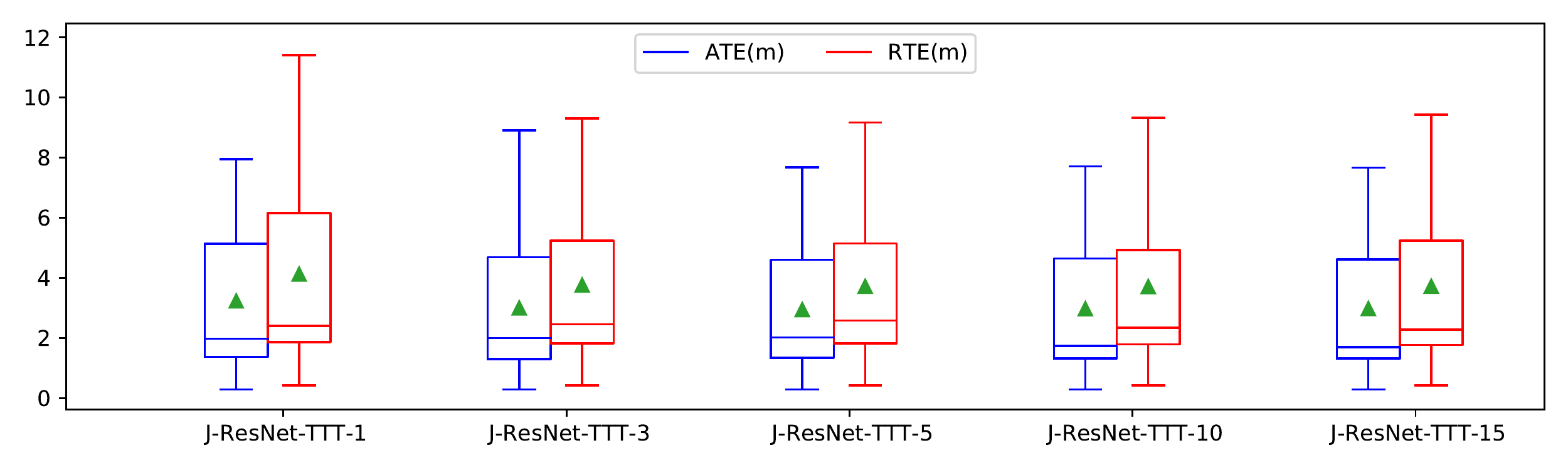}
	\vspace{-1ex}
	\caption{Impact of the number of TTT iterations on ATE and RTE}
	\label{fig:iter_oxiod}
\end{figure}

\begin{figure}
	\centering
	\includegraphics[width=8cm]{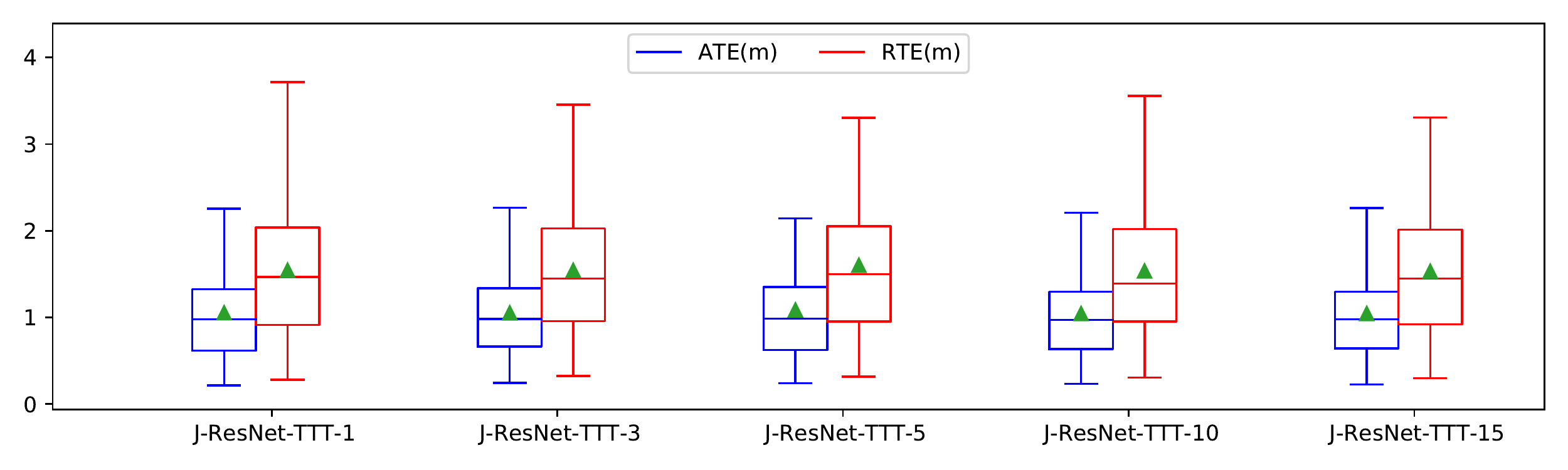}
	\vspace{-1ex}
	\caption{Impact of the number of TTT iterations on ATE and RTE}
	\label{fig:iter_ridi}
\end{figure}

\begin{figure*}[htb]
	\centering 
	\begin{subfigure}{0.99\textwidth}
		\includegraphics[width=\linewidth]{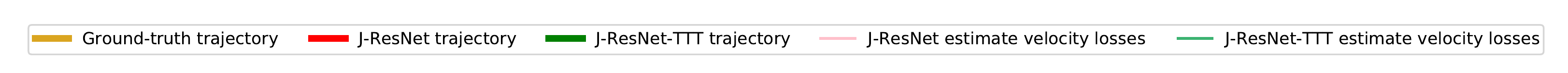}
		\vspace{-3ex}
		\label{fig:0}
	\end{subfigure}\hfil 

	\medskip
	\vspace{-2ex}
	\begin{subfigure}{0.33\textwidth}
		\includegraphics[width=\linewidth]{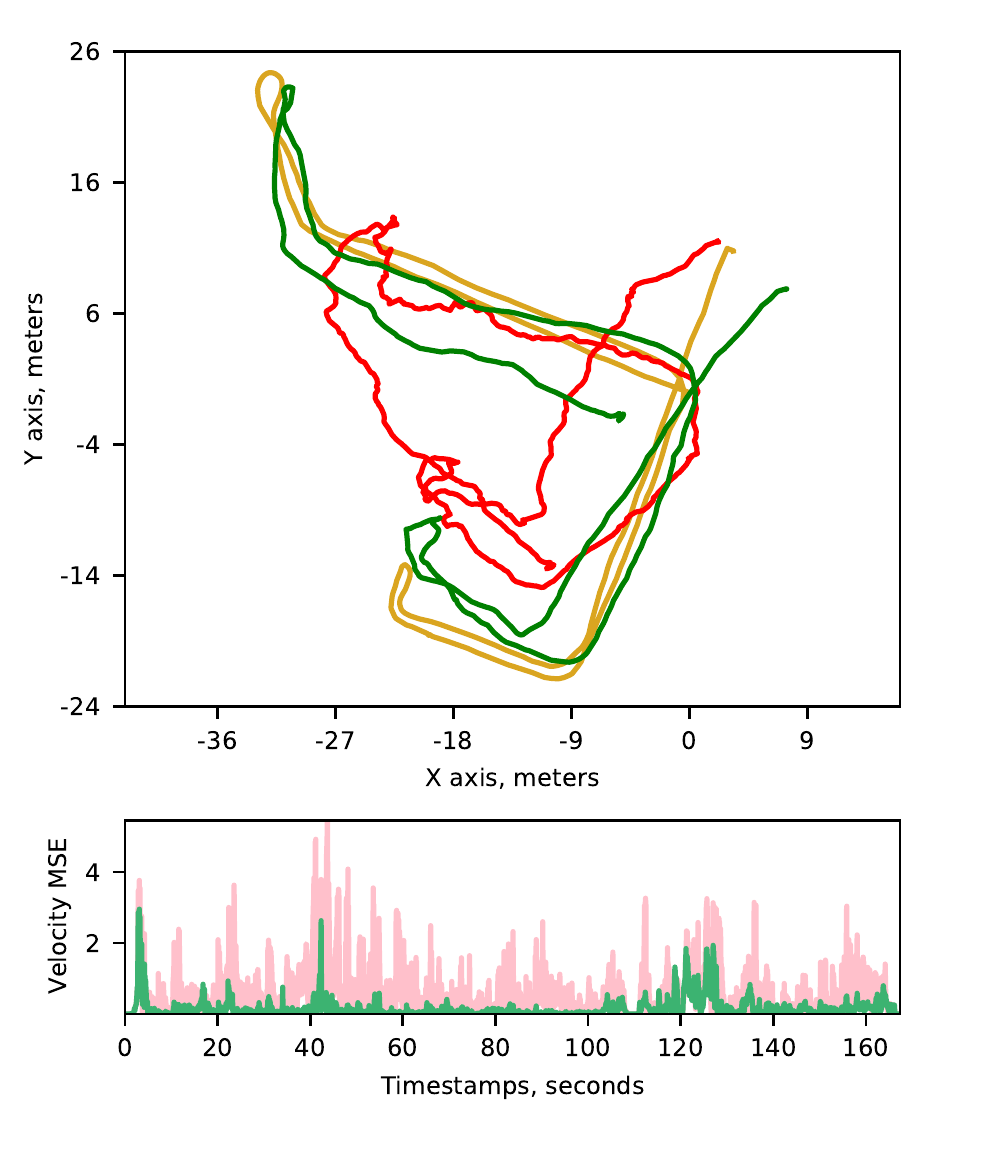}
		\vspace{-4.5ex}
		\caption{RIDI trajectory1}
		\label{fig:1}
	\end{subfigure}\hfil 
	\begin{subfigure}{0.33\textwidth}
		\includegraphics[width=\linewidth]{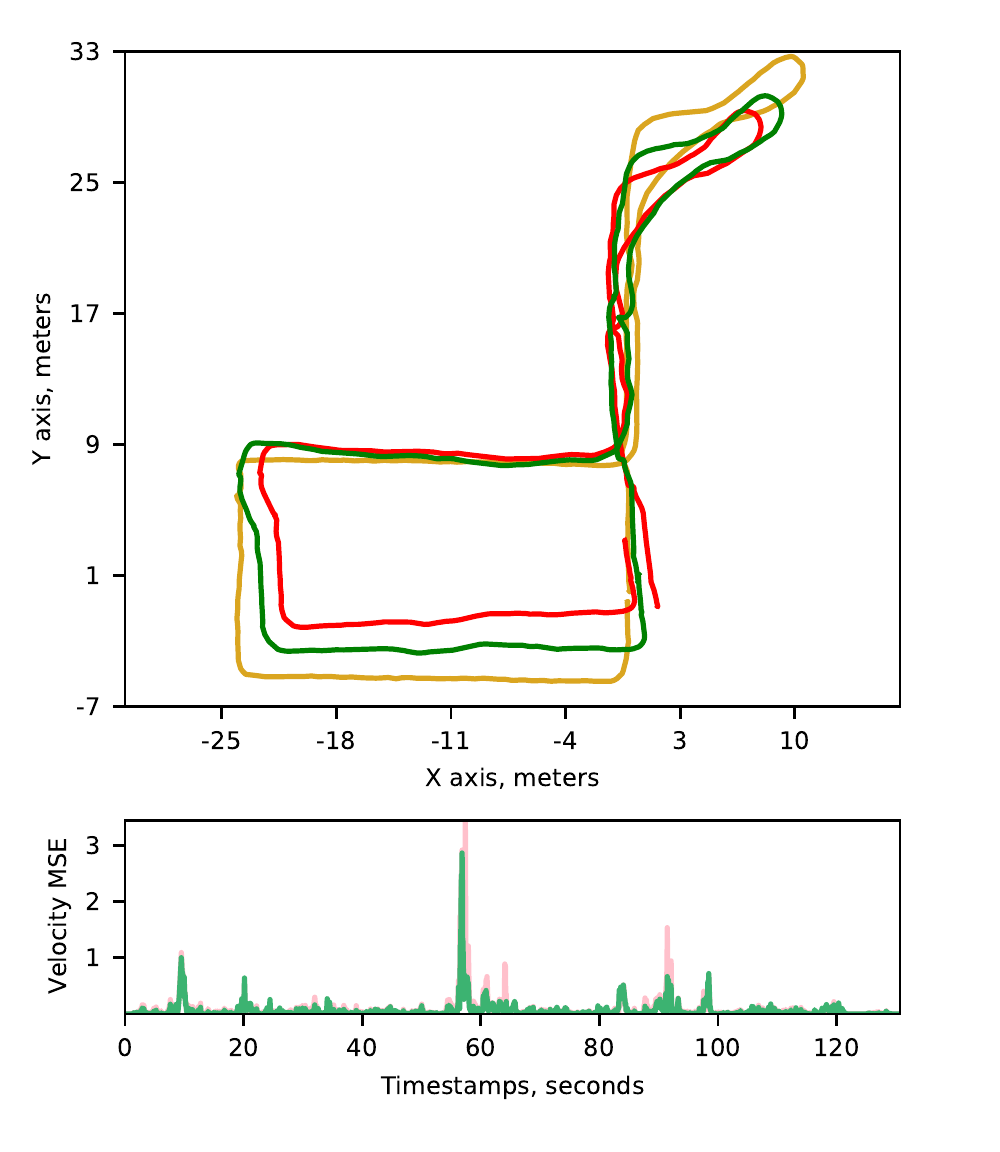}
		\vspace{-4.5ex}
		\caption{RIDI trajectory2}
		\label{fig:2}
	\end{subfigure}\hfil 
	\begin{subfigure}{0.33\textwidth}
		\includegraphics[width=\linewidth]{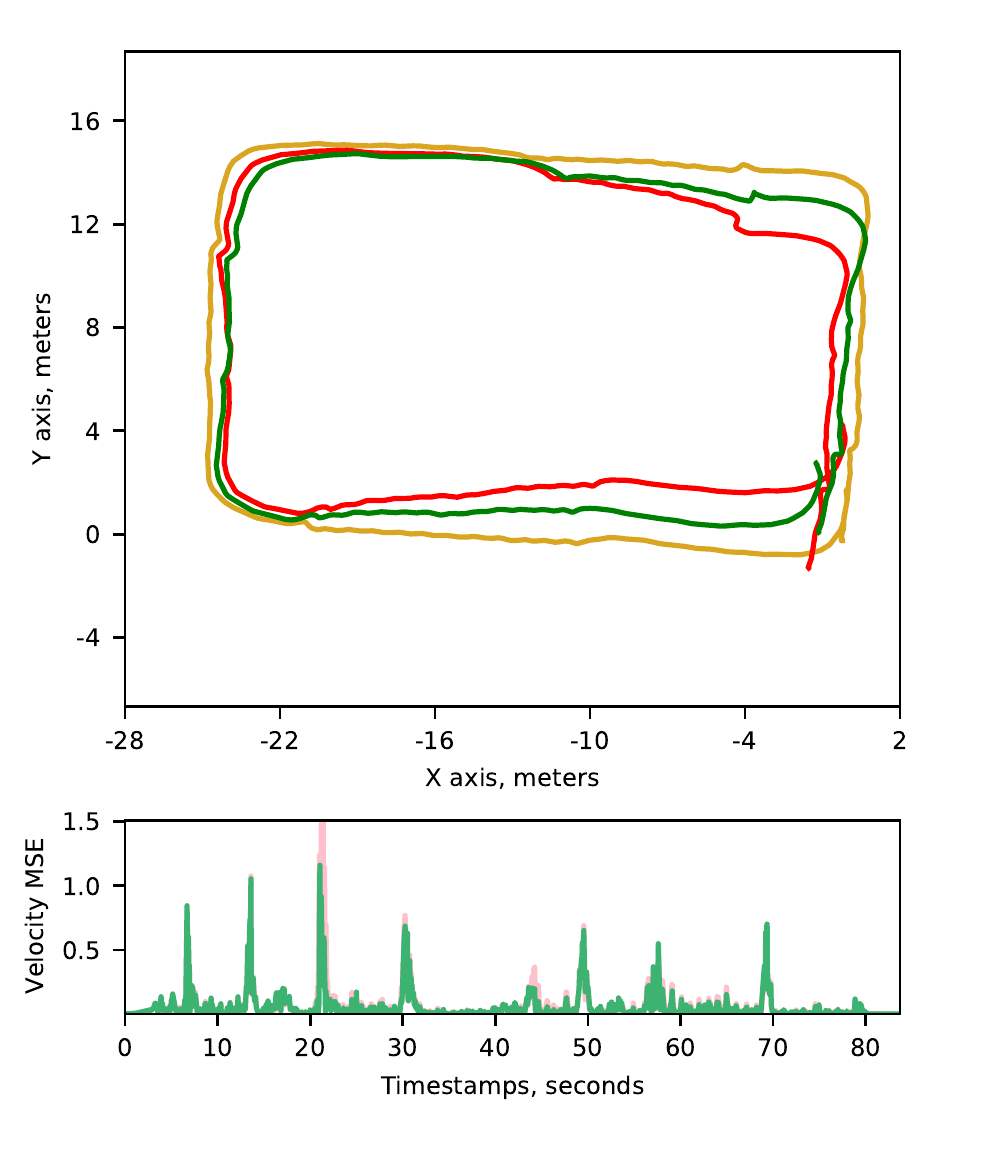}
		\vspace{-4.5ex}
		\caption{RIDI trajectory3}
		\label{fig:3}
	\end{subfigure}
	
	\medskip
	\begin{subfigure}{0.33\textwidth}
		\includegraphics[width=\linewidth]{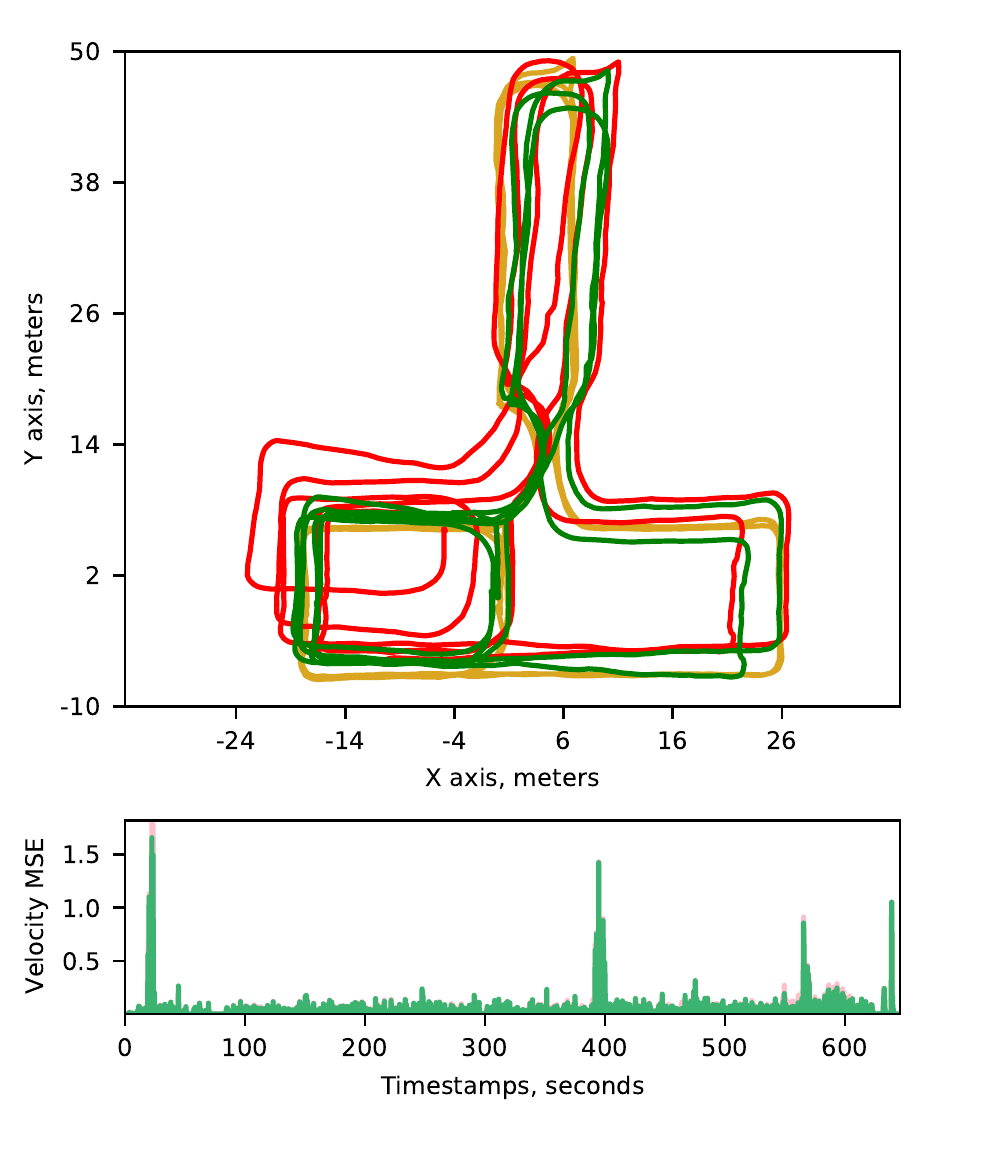}
		\vspace{-4.5ex}
		\caption{RoNIN trajectory1}
		\label{fig:4}
	\end{subfigure}\hfil 
	\begin{subfigure}{0.33\textwidth}
		\includegraphics[width=\linewidth]{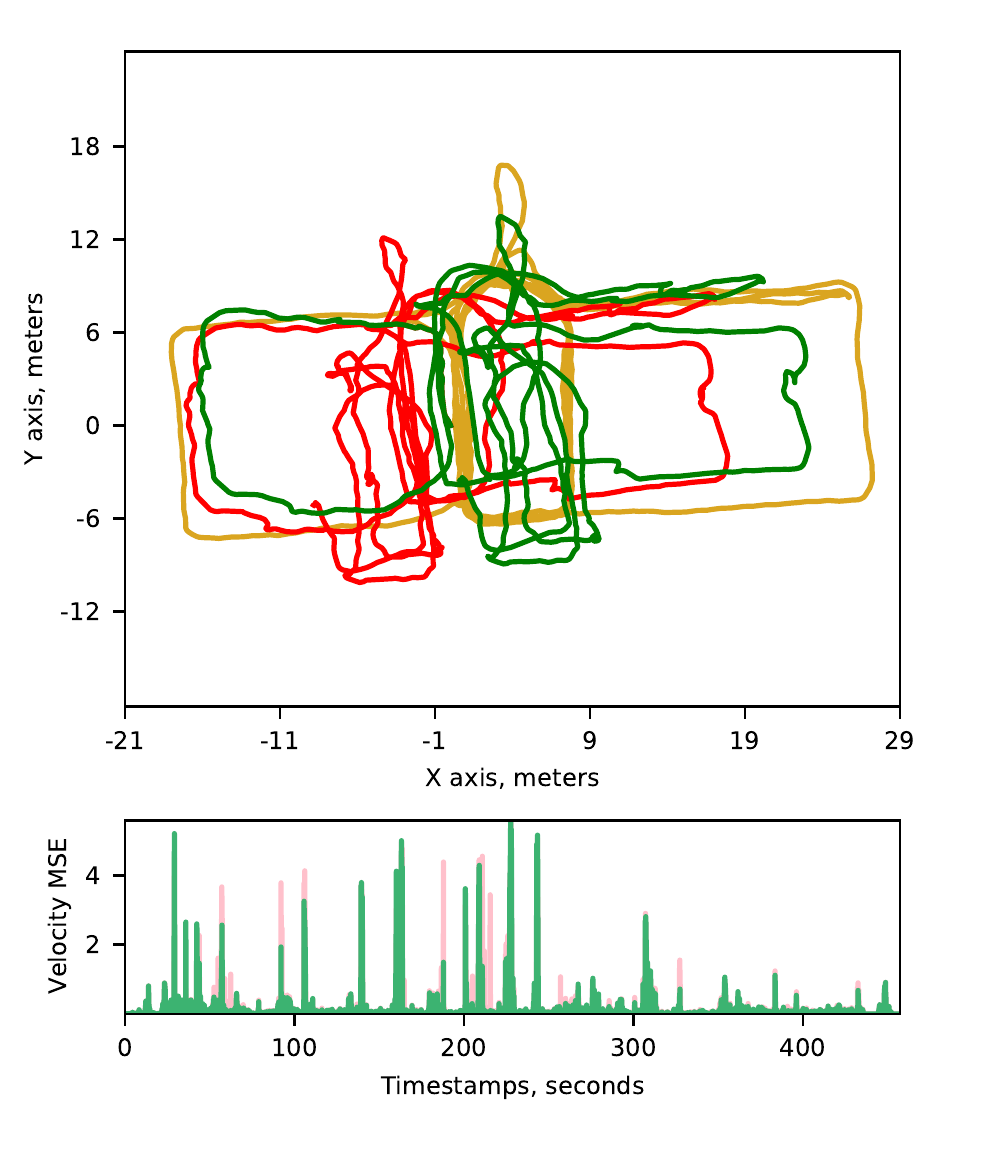}
		\vspace{-4.5ex}
		\caption{RoNIN trajectory2}
		\label{fig:5}
	\end{subfigure}\hfil 
	\begin{subfigure}{0.33\textwidth}
		\includegraphics[width=\linewidth]{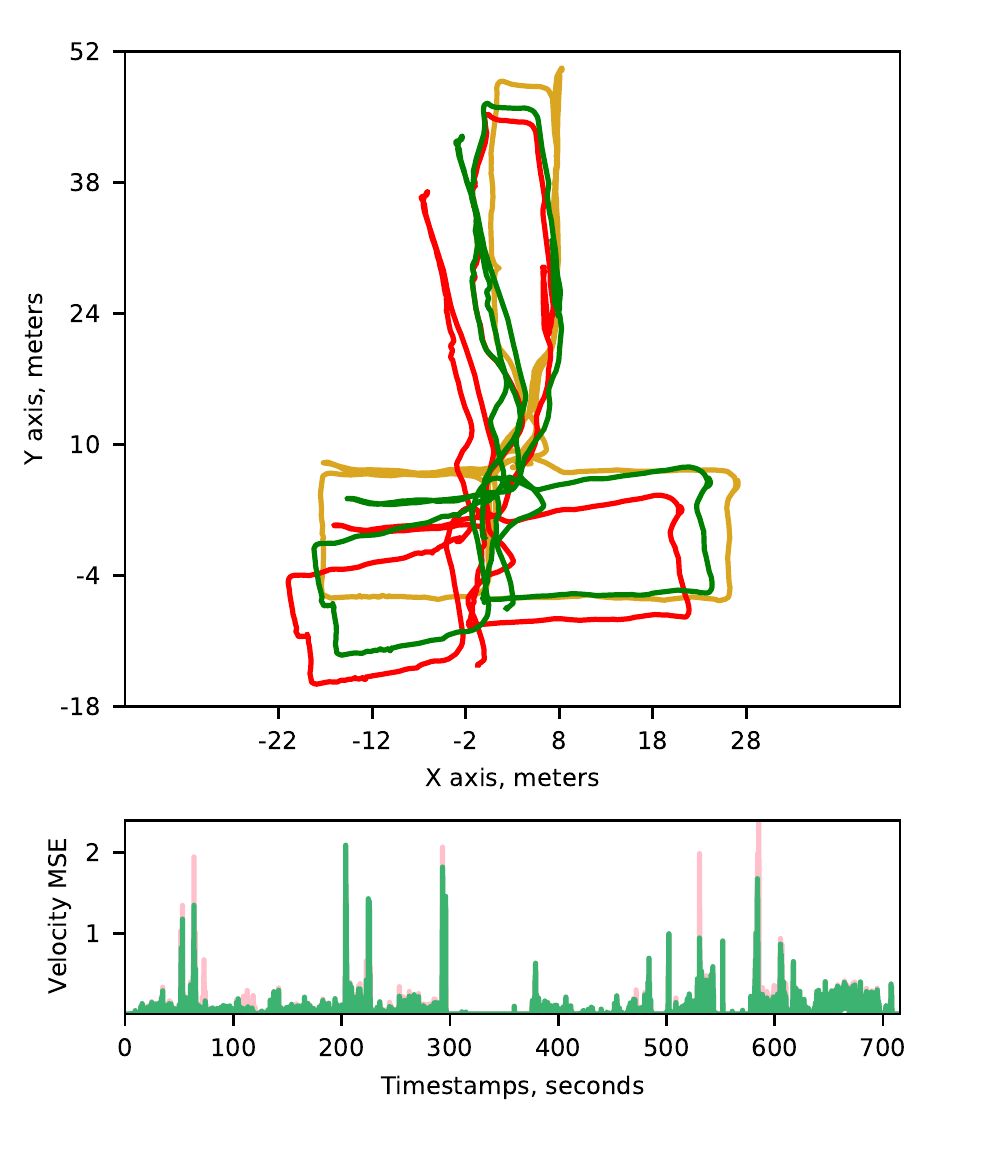}
		\vspace{-4.5ex}
		\caption{RoNIN trajectory3}
		\label{fig:6}
	\end{subfigure}

	\medskip
	\begin{subfigure}{0.33\textwidth}
		\includegraphics[width=\linewidth]{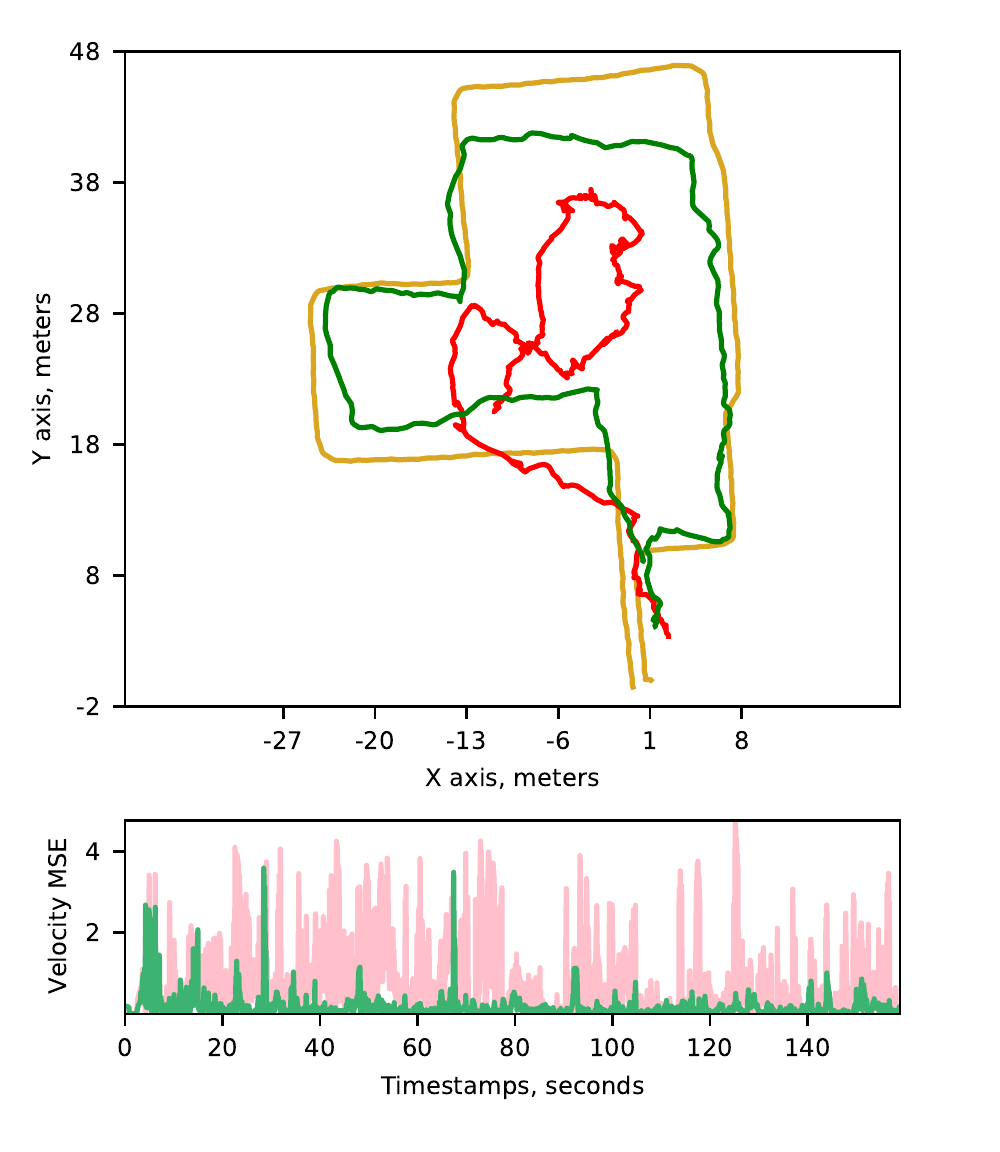}
		\vspace{-4.5ex}
		\caption{OXIOD trajectory1}
		\label{fig:4}
	\end{subfigure}\hfil 
	\begin{subfigure}{0.33\textwidth}
		\includegraphics[width=\linewidth]{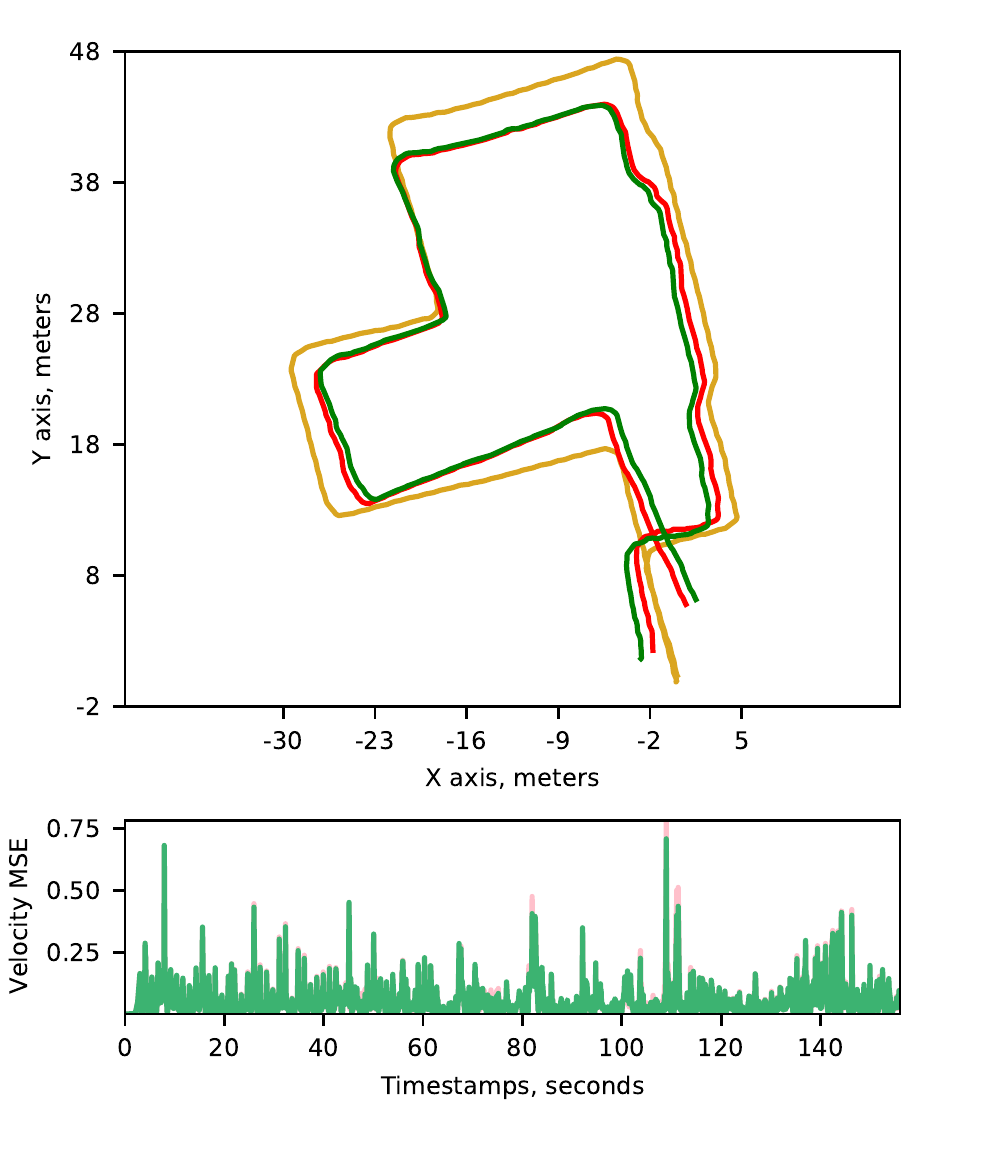}
		\vspace{-4.5ex}
		\caption{OXIOD trajectory2}
		\label{fig:5}
	\end{subfigure}\hfil 
	\begin{subfigure}{0.33\textwidth}
		\includegraphics[width=\linewidth]{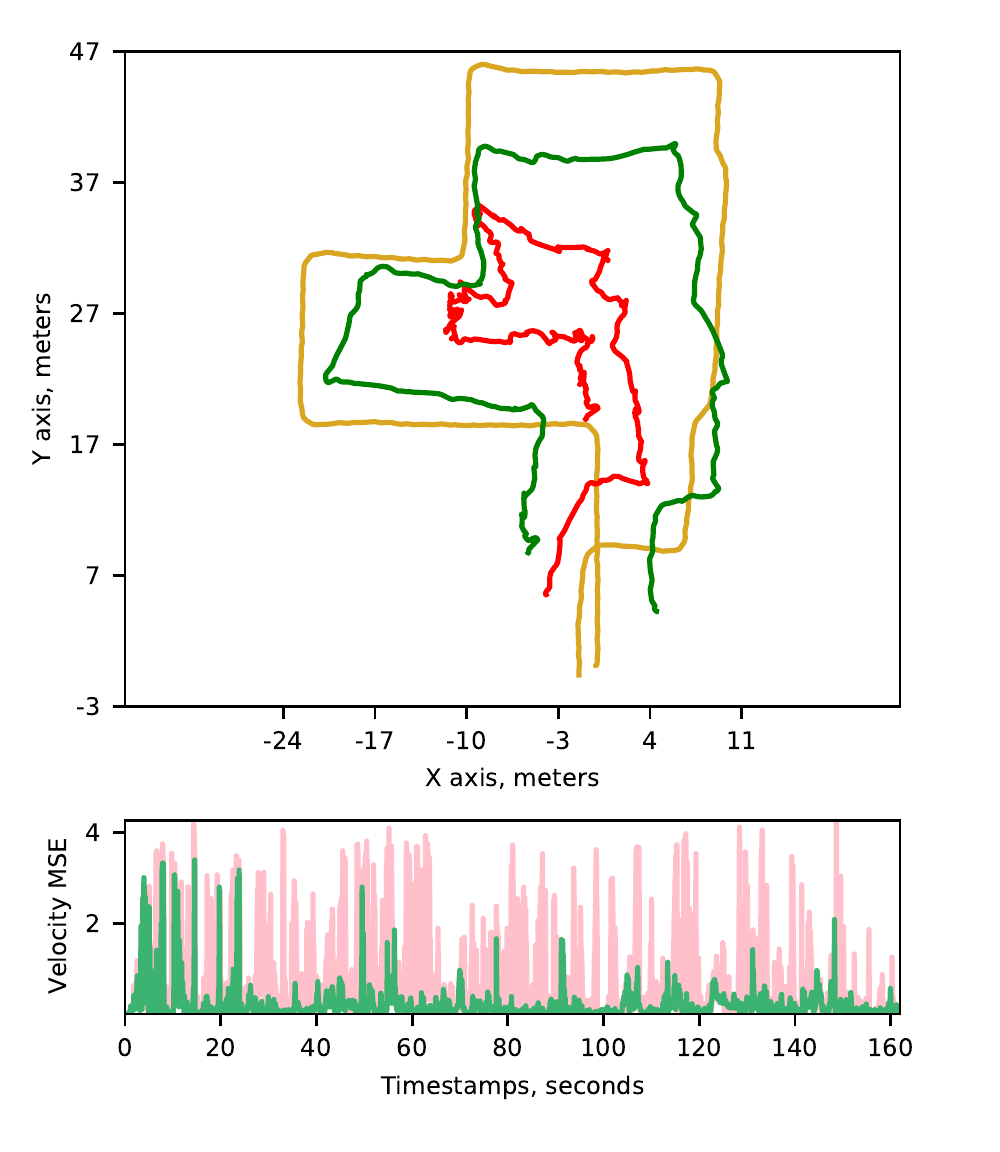}
		\vspace{-4.5ex}
		\caption{OXIOD trajectory3}
		\label{fig:6}
	\end{subfigure}
	\caption{Selected visualizations. We select 3 examples from each open sourced dataset and visualize trajectories of ground-truth, J-ResNet and J-ResNet-TTT, along with velocity losses comparison of J-ResNet and J-ResNet-TTT.}
	\label{fig:images}
\end{figure*}
\end{document}


	\title{RIO: Rotation-equivariance supervised learning of robust inertial odometry
	Supplementary Material}
	
	\author{Xiya Cao\textsuperscript{1}\thanks{\textsuperscript{1} denotes equal contribution.}
		\qquad
		Caifa Zhou\textsuperscript{1}
		\qquad
		Dandan Zeng
		\qquad
		Yongliang Wang\\
		Riemann lab, 2012 Laboratories, Huawei Technologies Co. Ltd\\
	}
	
	\maketitle

	\section*{Supplementary Materials}
	
	\textbf{Models performance v.s. size of training data} We trained models in joint-training setting with different size of training datasets. Denote the neural network which is provided and published by RONIN as 100\% B-ResNet since it is trained with the whole RONIN database. We trained models with 50\%, 30\% and 10\% data of the whole database in two ways as mentioned before. And evaluate their performance under different settings. \cref{fig:datasize} shows the comparison of different models. While \textit{B-ResNet} and \textit{J-ResNet} performances drop a lot as the training database becomes smaller, \textit{J-ResNet-TTT} with 30\% training database is still comparable to 100\% B-ResNet. However, \textit{J-ResNet-TTT} performance also drops a lot when using 10\% training databases.

	\textbf{Influence of updating iterations}  At test-time, model can be update multiple times with one batch of data. \cref{fig:iterations} shows results of one model with different updating iterations from 1 to 15. There are obvious improvements when increase iterations from 1 to 5. However, more than 5 updates do not show obvious advantages and the model performance even degrade a little when updating 15 times one batch. More iterations cause more time and computing resource consumption. Therefore, we recommend no more than 5 updates one batch during TTT.
	
	At test-time, model can be update multiple times with one batch of data. \cref{fig:iterations} shows results of one model with different updating iterations from 1 to 15. There are obvious improvements when increase iterations from 1 to 5. However, more than 5 updates do not show obvious advantages and the model performance even degrade a little when updating 15 times one batch. More iterations cause more time and computing resource consumption. Therefore, we recommend no more than 5 updates one batch during TTT.

	\textbf{Influence of updating iterations}  At test-time, model can be update multiple times with one batch of data. \cref{fig:iterations} shows results of one model with different updating iterations from 1 to 15. There are obvious improvements when increase iterations from 1 to 5. However, more than 5 updates do not show obvious advantages and the model performance even degrade a little when updating 15 times one batch. More iterations cause more time and computing resource consumption. Therefore, we recommend no more than 5 updates one batch during TTT.
	
	At test-time, model can be update multiple times with one batch of data. \cref{fig:iterations} shows results of one model with different updating iterations from 1 to 15. There are obvious improvements when increase iterations from 1 to 5. However, more than 5 updates do not show obvious advantages and the model performance even degrade a little when updating 15 times one batch. More iterations cause more time and computing resource consumption. Therefore, we recommend no more than 5 updates one batch during TTT.

	\begin{figure}
		\centering
		\includegraphics[width=8cm]{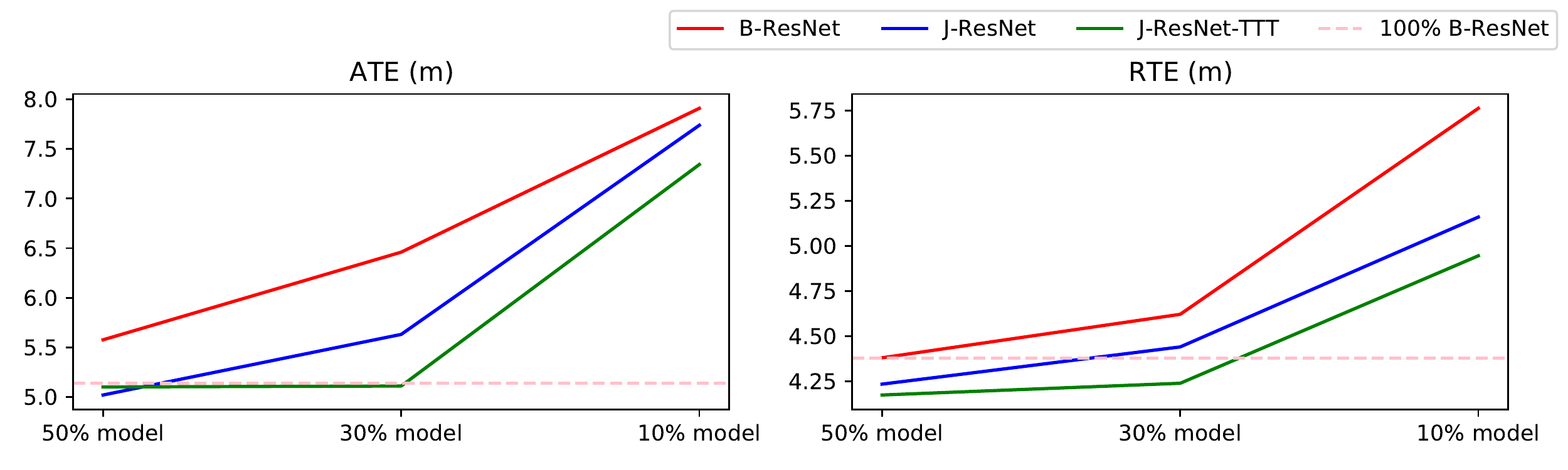}
		\vspace{-1ex}
		\caption{Impact of the number of TTT iterations on ATE and RTE}
		\label{fig:met_ronin}
	\end{figure}

	\begin{figure}
		\centering
		\includegraphics[width=8cm]{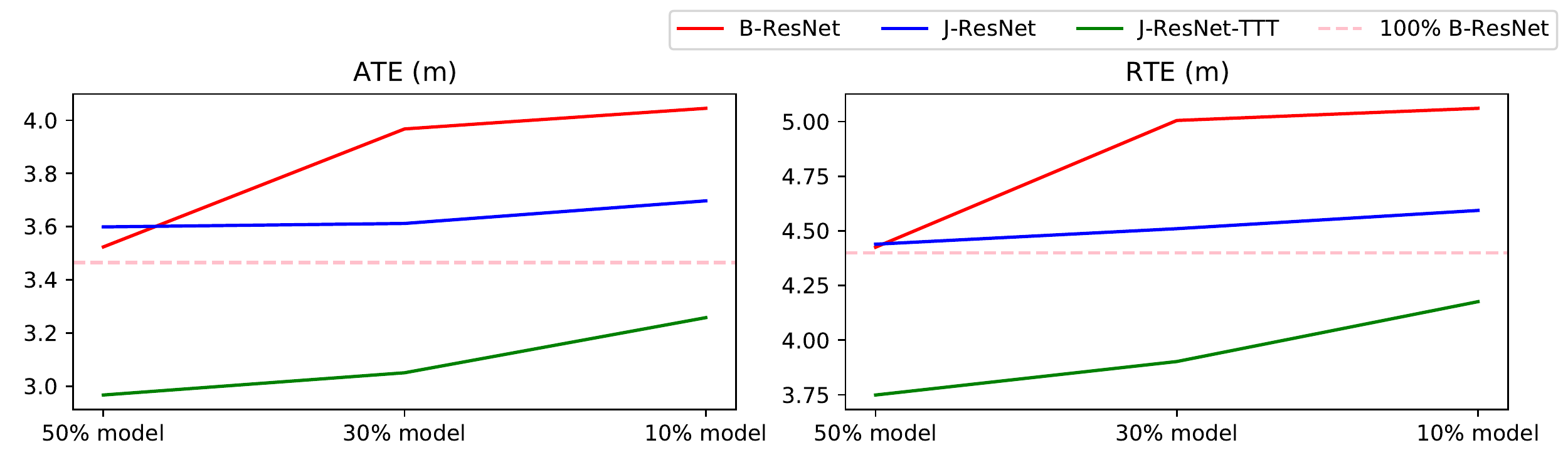}
		\vspace{-1ex}
		\caption{Impact of the number of TTT iterations on ATE and RTE}
		\label{fig:met_oxiod}
	\end{figure}

	\begin{figure}
		\centering
		\includegraphics[width=8cm]{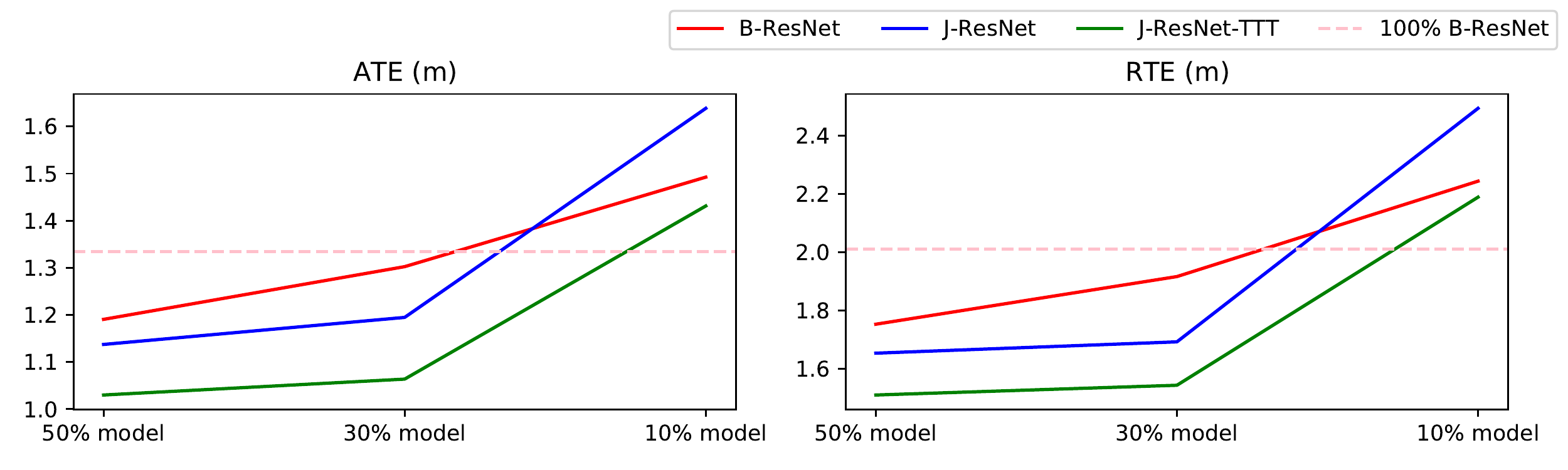}
		\vspace{-1ex}
		\caption{Impact of the number of TTT iterations on ATE and RTE}
		\label{fig:met_ridi}
	\end{figure}

	\begin{figure}
		\centering
		\includegraphics[width=8cm]{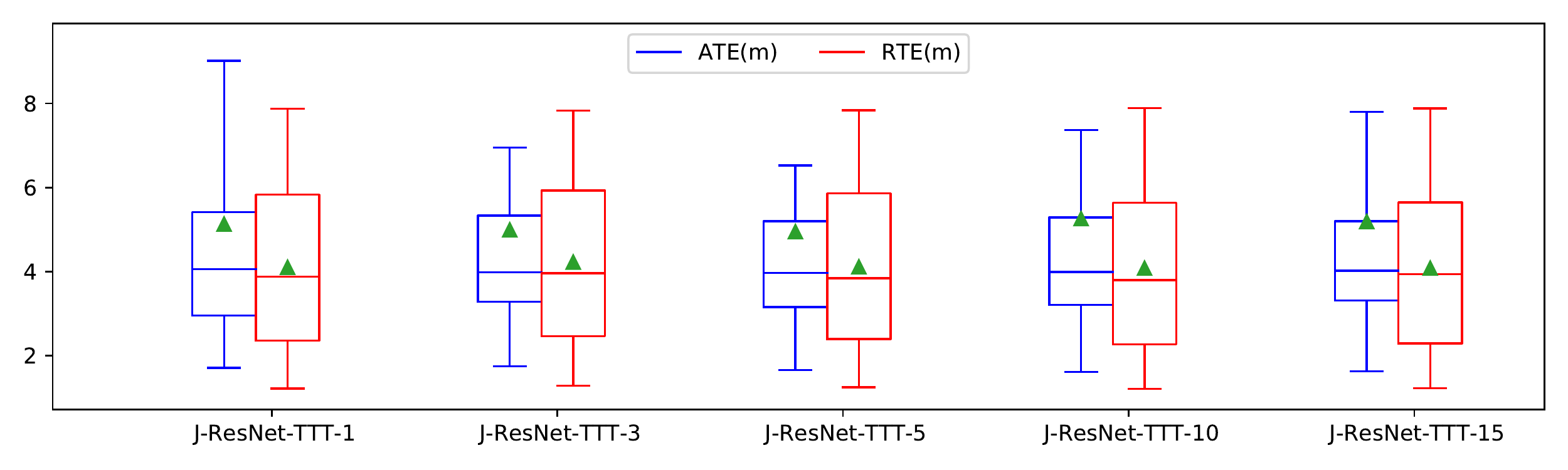}
		\vspace{-1ex}
		\caption{Impact of the number of TTT iterations on ATE and RTE}
		\label{fig:iter_ronin}
	\end{figure}

	\begin{figure}
		\centering
		\includegraphics[width=8cm]{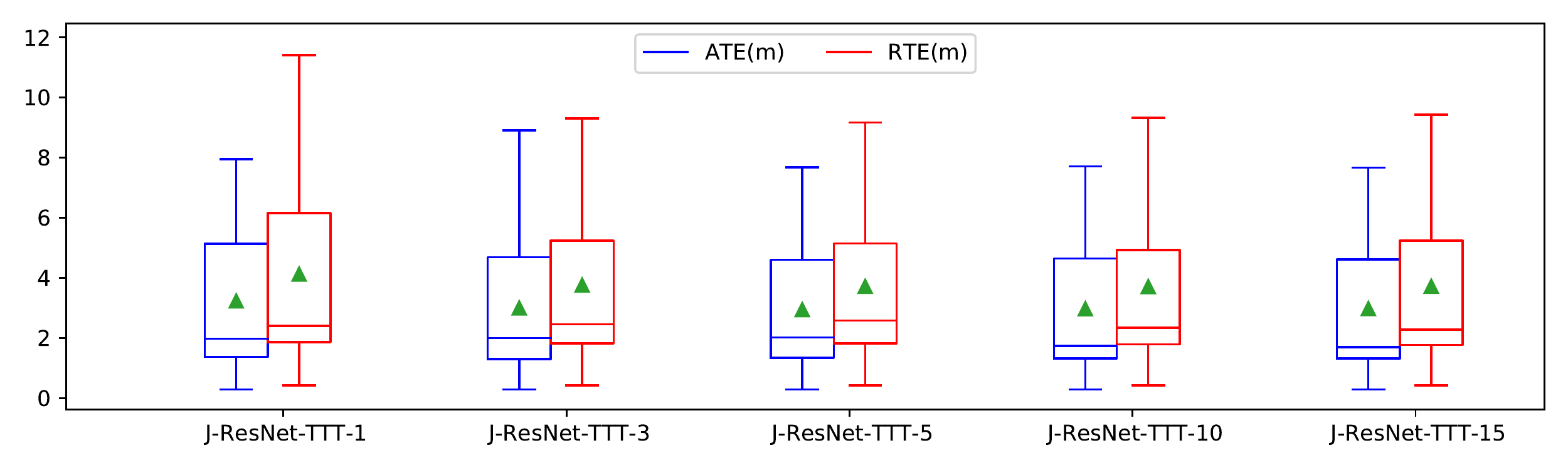}
		\vspace{-1ex}
		\caption{Impact of the number of TTT iterations on ATE and RTE}
		\label{fig:iter_oxiod}
	\end{figure}

	\begin{figure}
		\centering
		\includegraphics[width=8cm]{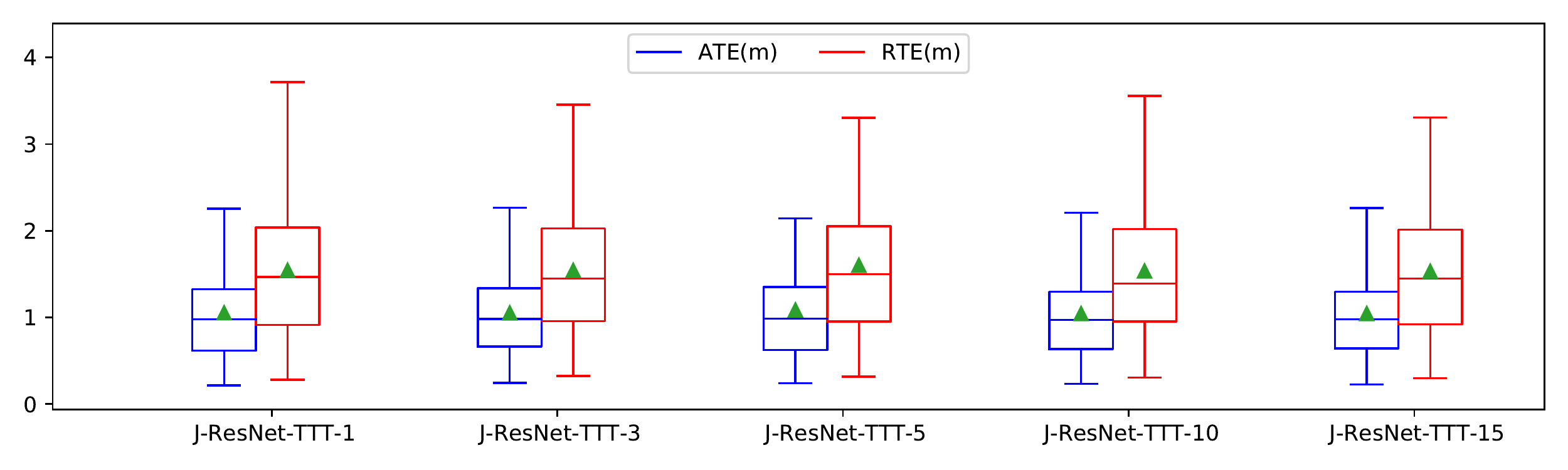}
		\vspace{-1ex}
		\caption{Impact of the number of TTT iterations on ATE and RTE}
		\label{fig:iter_ridi}
	\end{figure}

	\begin{figure*}[htb]
		\centering 
		\begin{subfigure}{0.99\textwidth}
			\includegraphics[width=\linewidth]{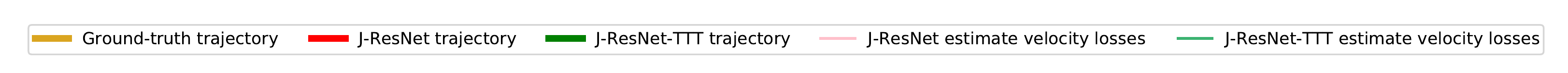}
			\vspace{-3ex}
			\label{fig:0}
		\end{subfigure}\hfil 

		\medskip
		\vspace{-2ex}
		\begin{subfigure}{0.33\textwidth}
			\includegraphics[width=\linewidth]{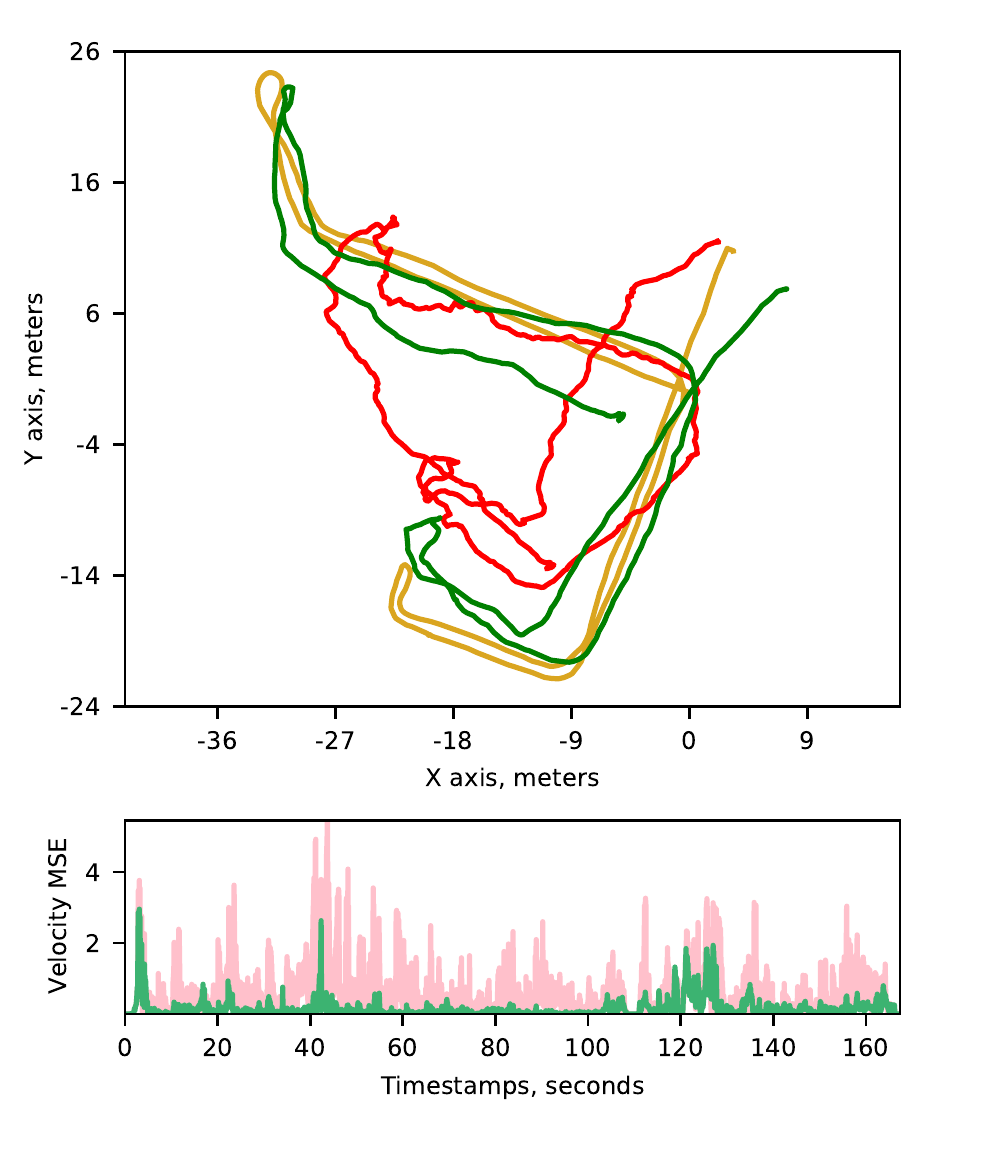}
			\vspace{-4.5ex}
			\caption{RIDI trajectory1}
			\label{fig:1}
		\end{subfigure}\hfil 
		\begin{subfigure}{0.33\textwidth}
			\includegraphics[width=\linewidth]{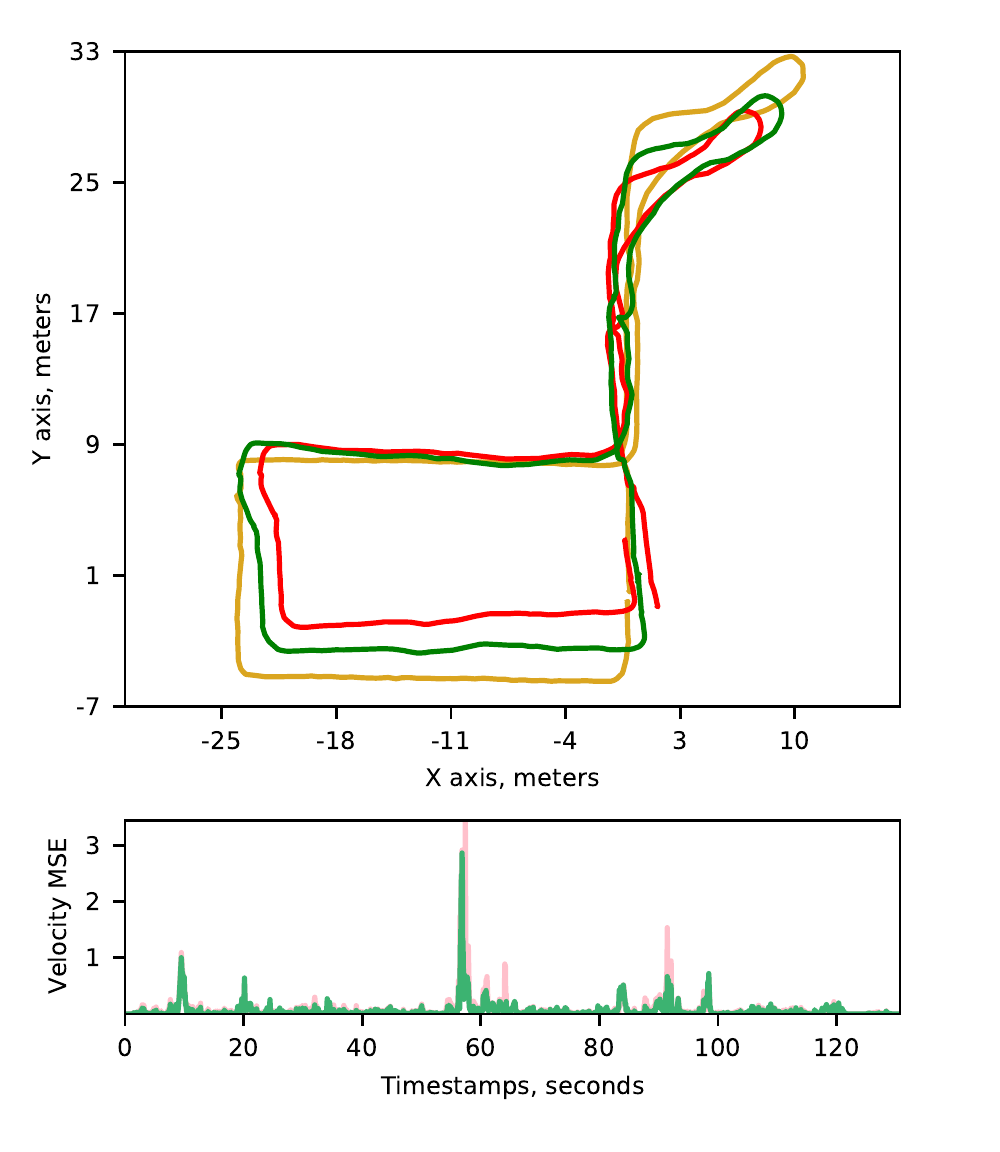}
			\vspace{-4.5ex}
			\caption{RIDI trajectory2}
			\label{fig:2}
		\end{subfigure}\hfil 
		\begin{subfigure}{0.33\textwidth}
			\includegraphics[width=\linewidth]{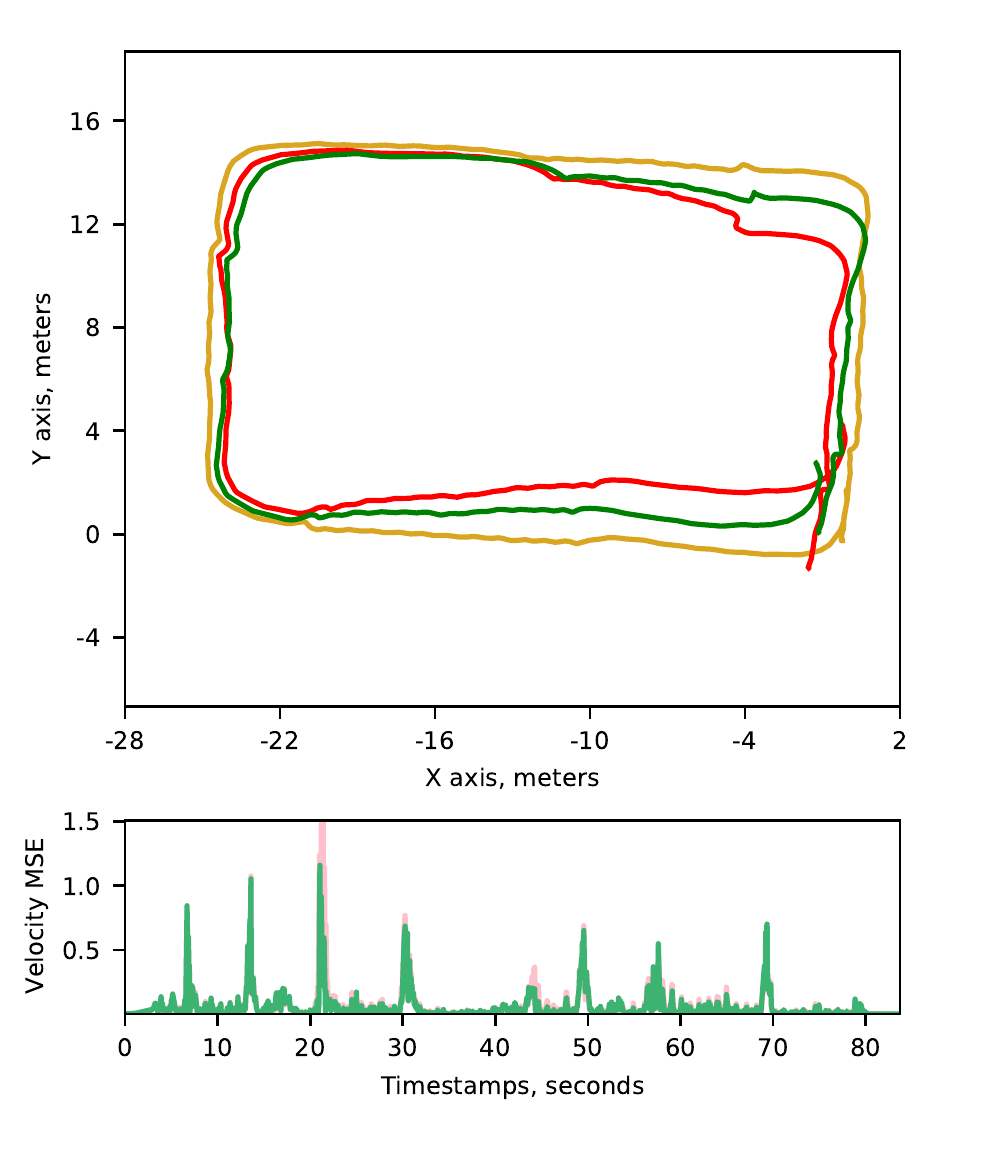}
			\vspace{-4.5ex}
			\caption{RIDI trajectory3}
			\label{fig:3}
		\end{subfigure}
		
		\medskip
		\begin{subfigure}{0.33\textwidth}
			\includegraphics[width=\linewidth]{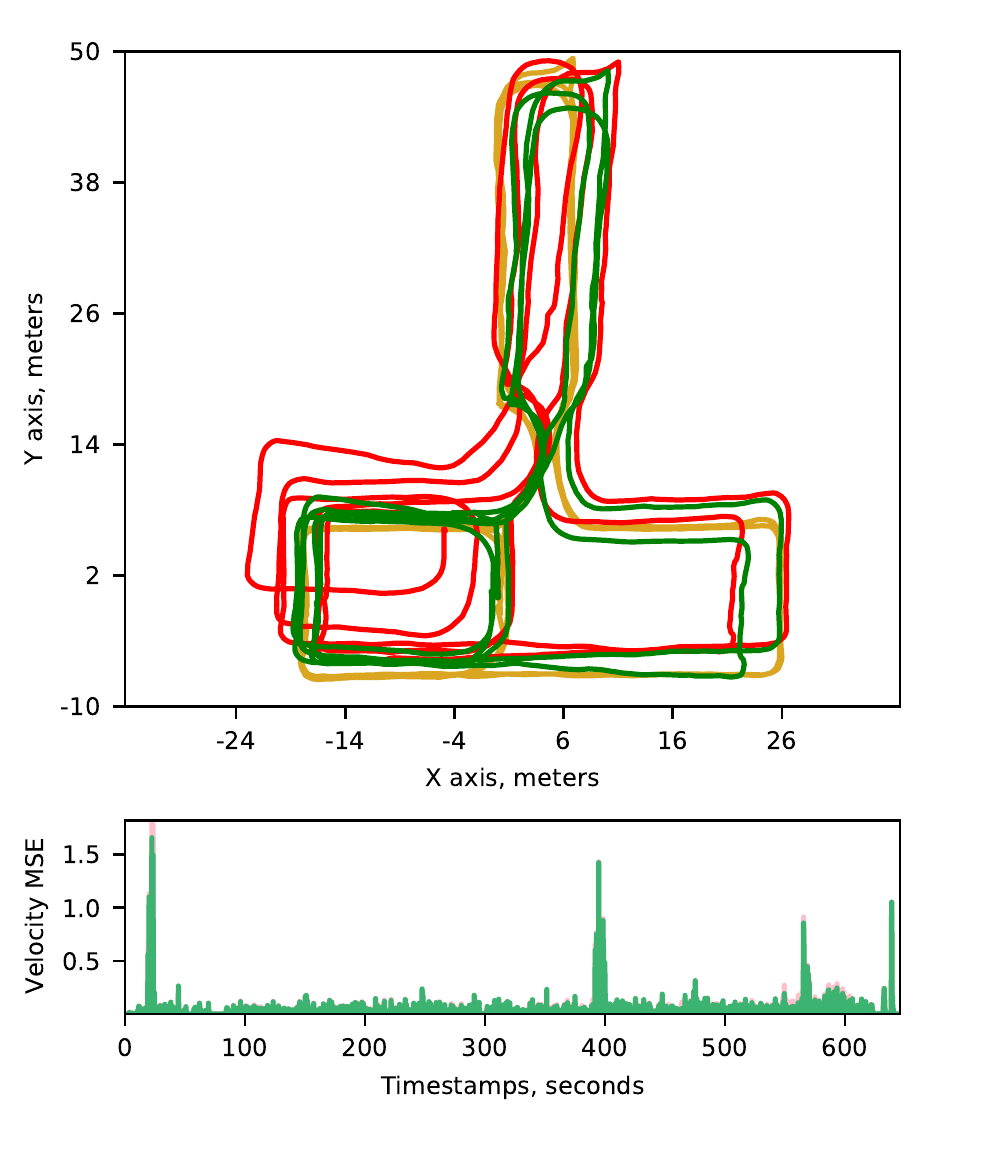}
			\vspace{-4.5ex}
			\caption{RoNIN trajectory1}
			\label{fig:4}
		\end{subfigure}\hfil 
		\begin{subfigure}{0.33\textwidth}
			\includegraphics[width=\linewidth]{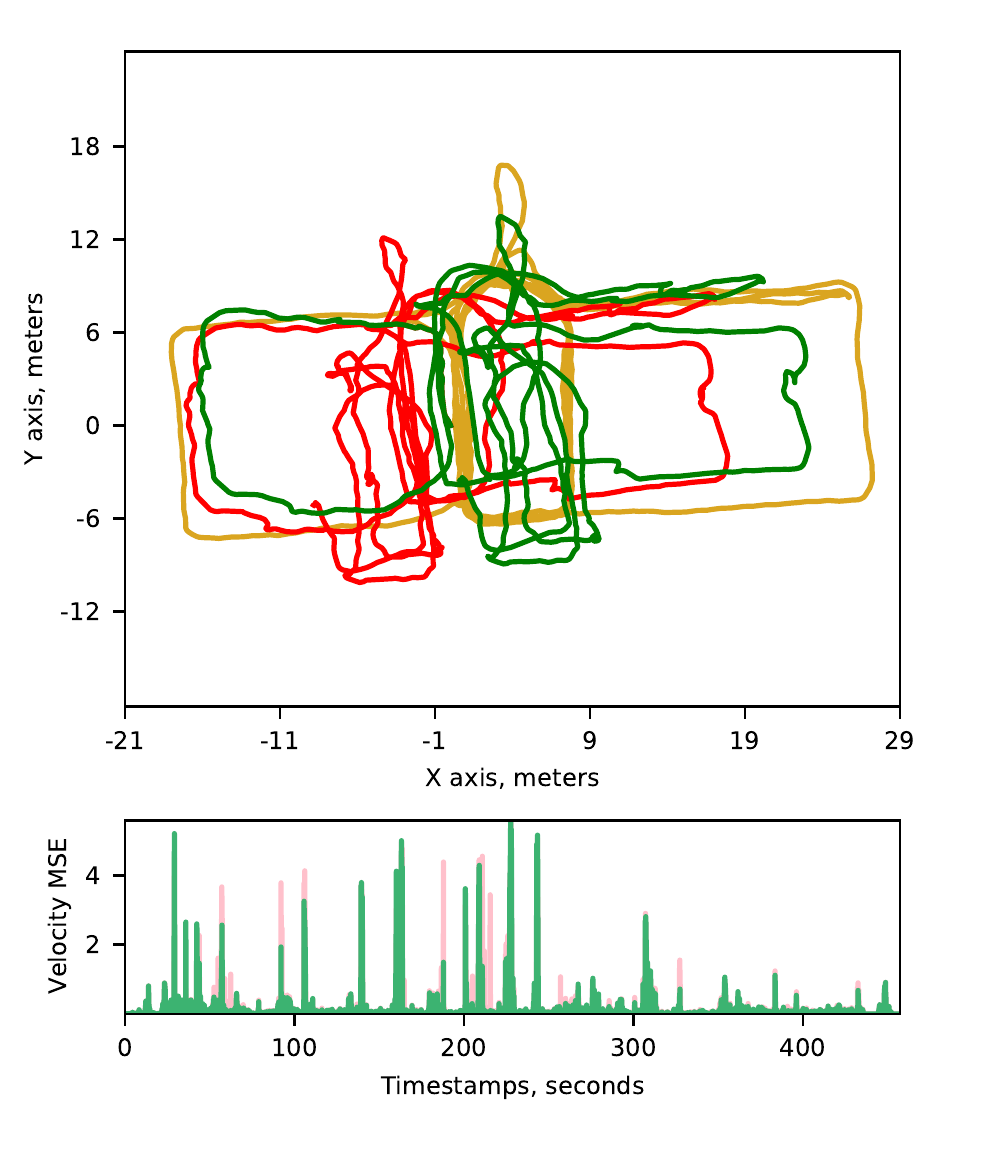}
			\vspace{-4.5ex}
			\caption{RoNIN trajectory2}
			\label{fig:5}
		\end{subfigure}\hfil 
		\begin{subfigure}{0.33\textwidth}
			\includegraphics[width=\linewidth]{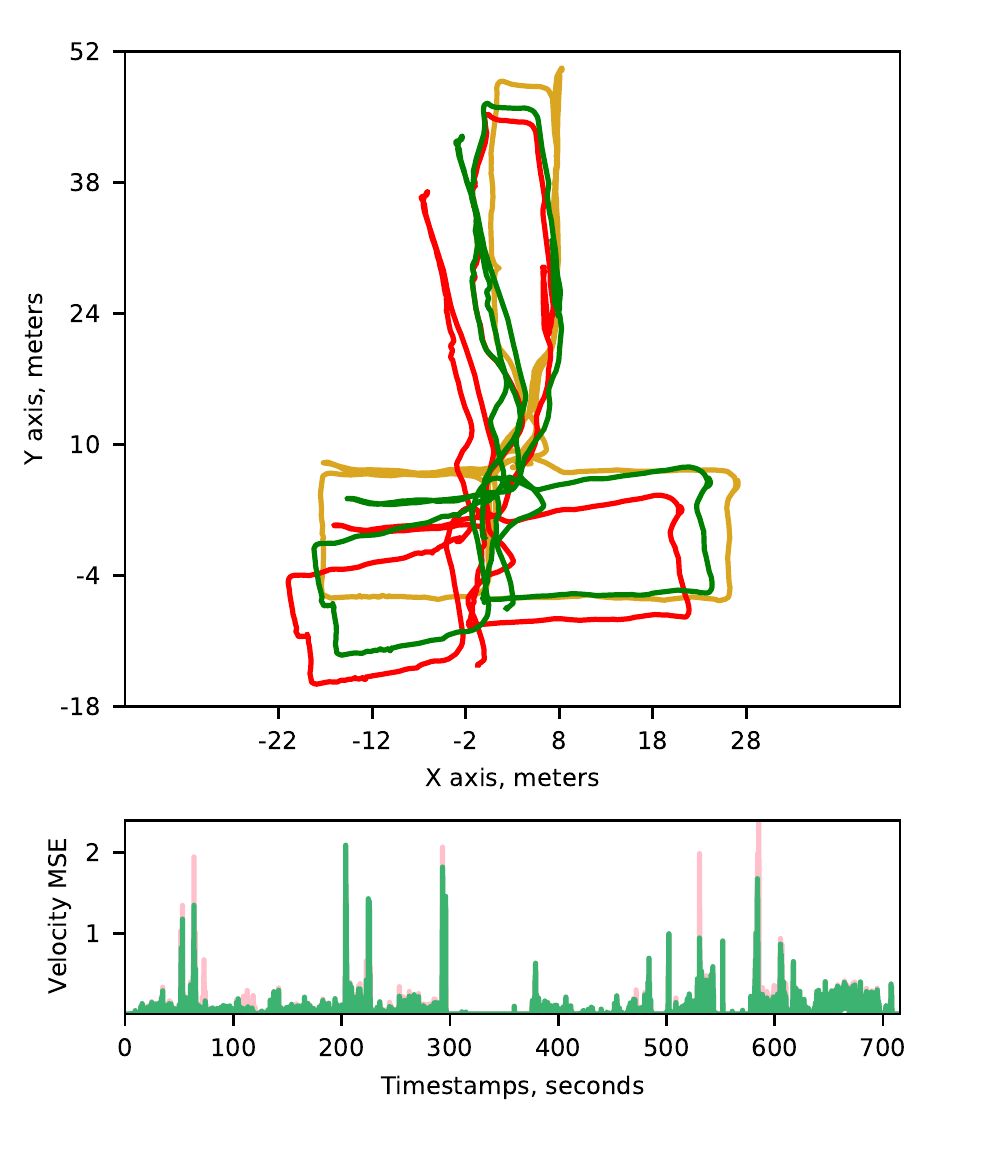}
			\vspace{-4.5ex}
			\caption{RoNIN trajectory3}
			\label{fig:6}
		\end{subfigure}

		\medskip
		\begin{subfigure}{0.33\textwidth}
			\includegraphics[width=\linewidth]{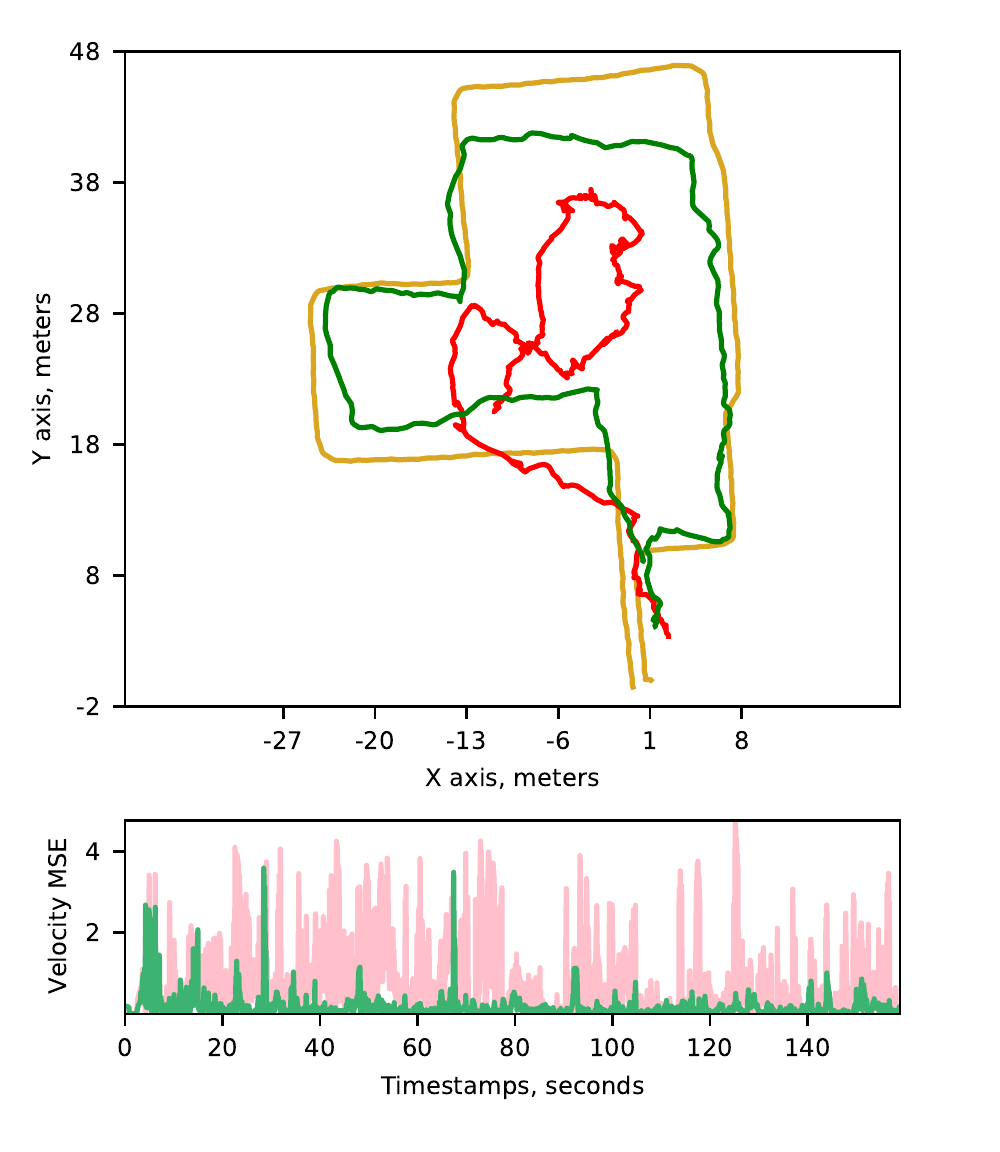}
			\vspace{-4.5ex}
			\caption{OXIOD trajectory1}
			\label{fig:4}
		\end{subfigure}\hfil 
		\begin{subfigure}{0.33\textwidth}
			\includegraphics[width=\linewidth]{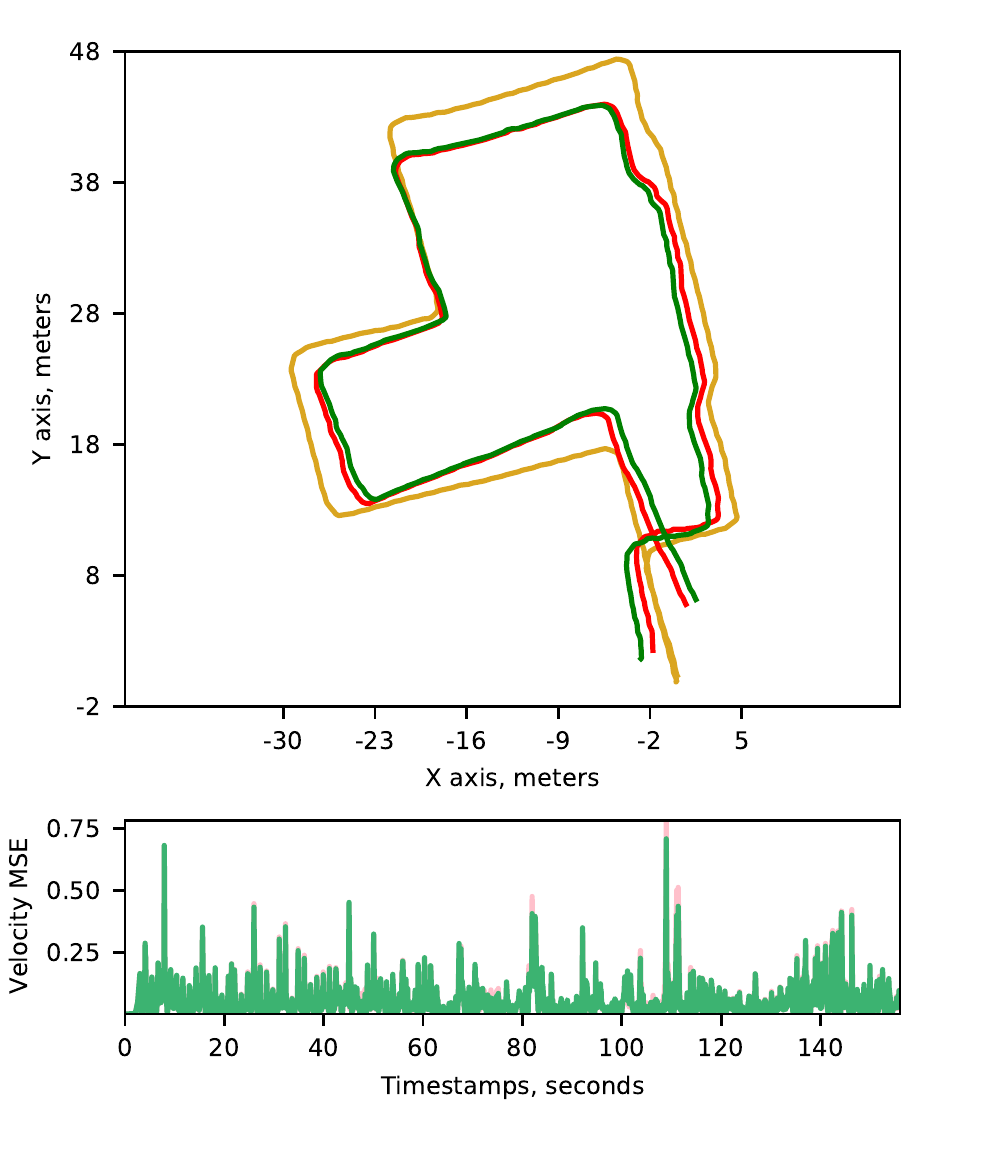}
			\vspace{-4.5ex}
			\caption{OXIOD trajectory2}
			\label{fig:5}
		\end{subfigure}\hfil 
		\begin{subfigure}{0.33\textwidth}
			\includegraphics[width=\linewidth]{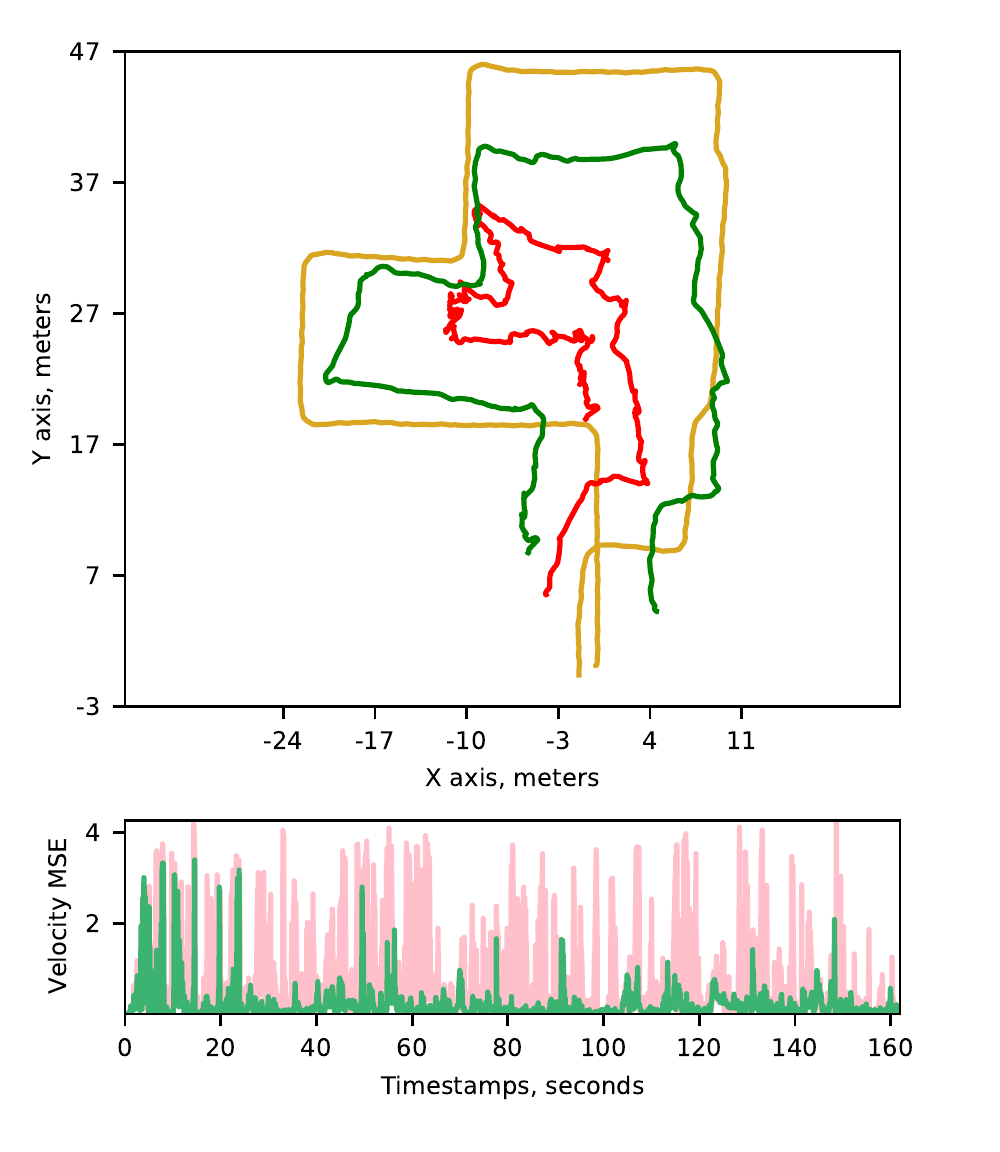}
			\vspace{-4.5ex}
			\caption{OXIOD trajectory3}
			\label{fig:6}
		\end{subfigure}
		\caption{Selected visualizations. We select 3 examples from each open sourced dataset and visualize trajectories of ground-truth, J-ResNet and J-ResNet-TTT, along with velocity losses comparison of J-ResNet and J-ResNet-TTT.}
		\label{fig:images}
	\end{figure*}